\pdfoutput=1

\documentclass[11pt]{article}
\usepackage[CJKbookmarks]{hyperref}  
\usepackage{EMNLP2023}

\usepackage{times}
\usepackage{latexsym}
\usepackage[T1]{fontenc}

\usepackage[utf8]{inputenc}

\usepackage{microtype}

\usepackage{inconsolata}

\usepackage{graphicx}
\usepackage{xcolor}
\usepackage{CJKutf8}
\usepackage{tabularx}
\usepackage{booktabs}
\usepackage{makecell}
\usepackage[misc]{ifsym}

\usepackage{bbding}
\usepackage{multirow}
\usepackage{arydshln}
\usepackage{subfigure}
\usepackage{caption}
\usepackage{ulem}
\usepackage{soul}
\usepackage{array,hhline}
\usepackage{amsmath}
\usepackage{bm}
\usepackage{ulem}
\usepackage{tikz}
\usepackage{pgf}
\usepackage{pgfplots}
\usetikzlibrary{
  arrows.meta,
  chains,
  matrix,
  positioning,
}

\usepackage{lipsum}

\makeatletter
\def\@currentColor{black}

\tikzset{
  set color/.code = { \gdef\@currentColor{#1} },
  box/.style = {
    draw,
    rounded corners=2mm,
    minimum width=1.8em,
    minimum height=1.8em,
    align=center,
    font=\color{black}\footnotesize\ttfamily,
    text height=1.0ex,
    text depth=.25ex,
  },
  box 1/.style = {box=#1, fill=orange!60},
  box 2/.style = {box=#1, fill=cyan!60},
  box 3/.style = {box=#1, fill=yellow!80!red!60},
  lower box/.style = {
    draw,
    fill=\@currentColor,
    minimum height=2.5em,
    inner xsep=2pt,
    font=\color{white}\ttfamily
  },
  matrix default/.style = {
    matrix of nodes
  },
  upper matrix/.style = {
    matrix default,
    column sep=15pt
  },
  lower matrix/.style = {
    matrix default,
    column sep=0pt,
    set color=green, 
    nodes=lower box
  },
  arrow/.style = {
    arrows={ -Triangle[angle=45:2.5pt 3] },
    line width=1pt
  },
  noarrowline/.style = {
    arrows={ -Triangle[angle=45:0pt 3] },
    line width=1pt
  },
  every on chain/.style = { 
    join=by arrow
  },
}
\makeatother

\newcommand{\zhsmall}[1]{\begin{CJK*}{UTF8}{gbsn}\small{#1}\end{CJK*}}

\definecolor{cmp1}{RGB}{0,129,218}
\definecolor{amp1}{RGB}{29,254,179}
\definecolor{ump1}{RGB}{62,254,94}
\definecolor{gmp1}{RGB}{194,254,0}
\definecolor{imp1}{RGB}{254,129,0}
\definecolor{xmp1}{RGB}{212,0,0}

\title{Lawyer LLaMA: Enhancing LLMs with Legal Knowledge}

\author{Quzhe Huang\thanks{\quad  Equal Contribution.} , Mingxu Tao$^*$, Chen Zhang$^*$, Zhenwei An$^*$, \\ 
\textbf{Cong Jiang}, \textbf{Zhibin Chen}, \textbf{Zirui Wu},  \Letter \textbf{Yansong Feng} \\
  Peking University \\
  {\tt \{huangquzhe,thomastao,zhangch,anzhenwei\}@pku.edu.cn} \\
  {\tt fengyansong@pku.edu.cn} \\}

\begin{document}
\maketitle
\begin{abstract}
Large Language Models (LLMs), like LLaMA, have exhibited remarkable performance across various tasks. Nevertheless, when deployed to specific domains such as law or medicine, the models still confront the challenge of a deficiency in domain-specific knowledge and an inadequate capability to leverage that knowledge to resolve domain-related problems. In this paper, we propose a new framework to adapt LLMs to specific domains and build Lawyer LLaMA, a legal domain LLM, based on this framework. Specifically, we inject domain knowledge during the continual training stage and teach the model to learn professional skills using properly designed supervised fine-tuning tasks.  Moreover, to alleviate the hallucination problem during the model's generation, we add a retrieval module and extract relevant legal articles before the model answers any queries. When learning domain-specific skills, we find that experts' experience is much more useful than experiences distilled from ChatGPT, where hundreds of expert-written data outperform tens of thousands of ChatGPT-generated ones. We will release our model and data.

\end{abstract}

\section{Introduction}

When enhancing open-source large language models with instruction tuning, models such as Vicunna~\cite{vicuna2023} and ChatGLM~\cite{zeng2023glm-130b} have achieved remarkable performance in various general tasks~\cite{zheng2023judging,open-llm-leaderboard,alpaca_eval}. 

\begin{figure}[t]
\centering
\includegraphics[scale=0.185]{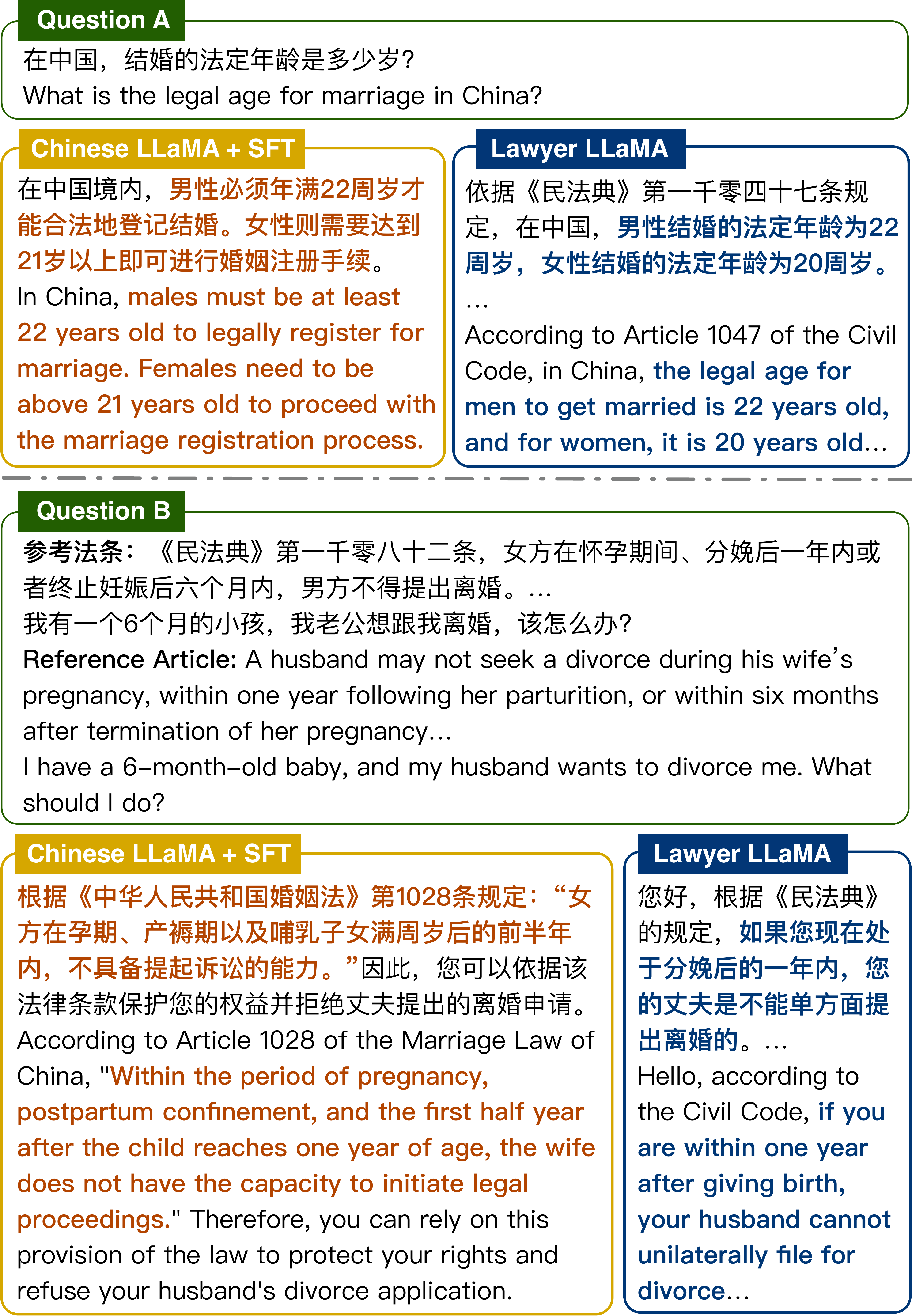}
\caption{Comparison between LLaMA continually trained on general Chinese corpus and instructions (Chinese LLaMA + SFT) and our Lawyer LLaMA, which is specially designed for the legal domain. It can be seen that Chinese LLaMA lacks domain-specific knowledge, e.g., \textit{legal age of marriage}. Even if the necessary knowledge, i.e.,  the specific legal article is given, Chinese LLaMA still cannot use it to solve practical problems. }
\label{fig:intro_example}
\vskip -0.2in
\end{figure}

However, due to the lack of data  and improper training paradigm, adapting large language models~(LLMs) to specific domains like the legal domain remains a challenging task. Specifically, there are two major issues need to solve:

\paragraph{Lacking domain knowledge} Current LLMs are primarily trained on general corpora such as C4~\cite{t5-c4} and Wikipedia\footnote{\url{https://dumps.wikimedia.org/enwiki/latest/}},
with limited exposure to domain-specific resources. Therefore, they lack the necessary knowledge required for specific domains. Figure~\ref{fig:intro_example} shows the responses to some legal questions from two LLMs, one is trained on the general corpus, Chinese-LLaMA~\cite{chinese-llama-alpaca}, and the other is our Lawyer LLaMA which is designed to solve legal issues. For Question A, the right answer is what Lawyer LLaMA provides, that the legal age of marriage is 22 for men and 20 for women. Obviously, Chinese-LLaMA lacks such domain-specific knowledge and incorrectly claims that women need to be 21 years old for marriage.
 
\paragraph{Unable to solve problems by using domain-specific knowledge} The strategy to analyze and solve domain-specific tasks might diverge significantly from what the model has obtained within the general domain. Even if the model could access all the knowledge within a specific domain, applying them to solve practical problems remains a formidable undertaking.  For example, in Question B, even if the necessary legal article is provided, Chinese-LLaMA cannot understand such knowledge. It claims "the \textit{wife} does not have the capacity to initiate legal proceedings", whereas the original legal article says the \textit{husband}. Moreover, Chinese-LLaMA directly provides the results, without analyzing the situation according to the legal article as Lawyer LLaMA does, which makes its results unreliable.

To alleviate the above problems, we propose a new framework to adapt LLMs to specific domains. It takes three steps to train a reliable domain-specific LLM and we illustrate our framework in the legal domain:

\paragraph{Inject domain-specific knowledge with diverse sources} We collect a large amount of raw  text in the legal domain, such as legal articles, judicial interpretations, and judicial documents of the People's Court of China. We then apply continual pre-training to help the model obtain legal knowledge.

\paragraph{Learn professional skills from experts} To solve practical problems, domain experts will analyze different concepts, and map abstract concepts and theories to specific scenarios. We collect solutions to practical problems by experts 
and use them to teach the model how to solve domain-specific tasks with proper knowledge. If domain experts are not available, powerful LLMs, such as ChatGPT, could be a substitute. However, for learning professional skills, hundreds of expert-written solutions are more valuable than tens of thousands of ChatGPT-generated ones, due to the difference in quality. This data is used for supervised fine-tuning(SFT), similar to Alpaca~\cite{alpaca}.

\paragraph{Augment with external knowledge and filter the irrelevant out} To alleviate the hallucination problem~\cite{Hallucination} and generate more reliable responses, we additionally introduce an information retrieval module. In the legal domain, legal articles are retrieved and serve as external knowledge to answer the queries of clients. We notice that not all retrieved knowledge is useful for a certain query. We thus design a special mechanism during training to teach the model to filter out irrelevant information.

In this paper, we propose a framework to adapt LLMs to specific domains. We apply our framework in the legal domain and build Lawyer LLaMA, an LLM specifically designed to solve legal problems. We find that expert experience is very efficient and effective when teaching the model to use domain knowledge. It might not be the best choice to  distill professional skills totally from ChatGPT without involving experts' efforts, as ~\citet{Visual-Med-Alpaca} do. Another interesting finding is that the model tends to utilize all the additional knowledge provided in the input, even if that knowledge is not helpful in answering the current question. Considering that augmenting LLMs with external knowledge is becoming increasingly popular in domain adaptation~\cite{yunxiang2023chatdoctor,xiong2023doctorglm}, we argue that such a technique should be used with caution and we should make sure that the models are able to filter out irrelevant information.

\section{Data Collection}
\subsection{Chinese Legal Corpus}
\label{sec:zh_legal_corpus}
Previous work~\cite{biobert} reveals that language models can leverage knowledge learned from a domain-specific corpus and that such models can handle tasks in the corresponding domain better than the models pre-trained only on the general corpus. To augment our model with Chinese legal knowledge, we collect texts from the websites of China Courts, including judgment documents, legal articles, judicial interpretations, court news, and various articles for law popularization.

Previous work~\cite{chen-etal-2020-recall} also reveals that adapting a language model to a new domain might cause catastrophic forgetting of its general knowledge. To alleviate forgetting of general knowledge, we sample Chinese texts from WuDaoCorpora~\cite{wudao-yuan-et-al} and CLUECorpus2020~\cite{xu2020cluecorpus2020} and English texts from C4~\cite{t5-c4} for sparse episodic replay~\cite{d2019episodic,tao2023can}.

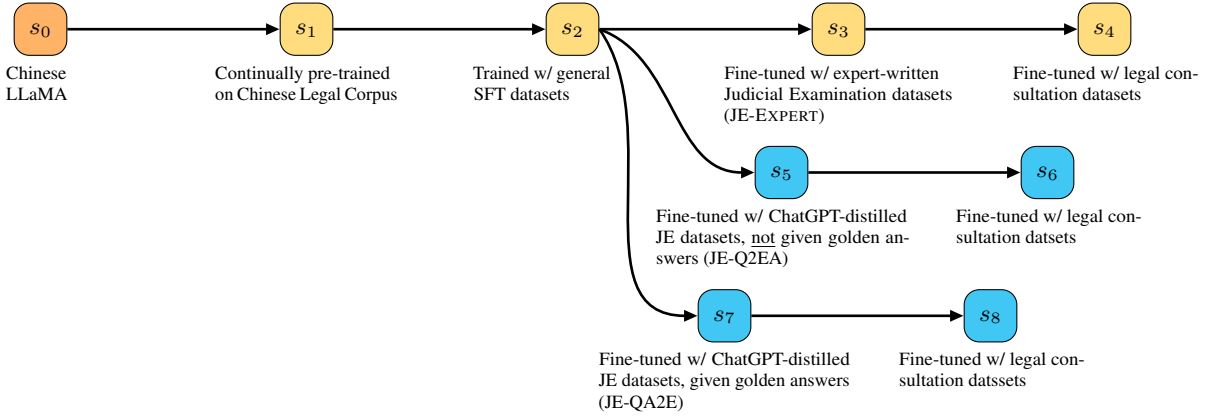
\begin{figure*}[ht]
\centering
\begin{tikzpicture}
  \node[box 1, label=below:{\parbox{0.9cm}{\scriptsize Chinese\\ LLaMA}}] (0) at (0,0){$s_0$};
  \node[box 3, label={[xshift=0cm]below:{\parbox{2.4cm}{\scriptsize Continually pre-trained\\ on Chinese Legal Corpus}}}] (1) at (3.5,0){$s_1$};
  \node[box 3, label={[xshift=-0.4cm]below:{\parbox{1.8cm}{\scriptsize Trained w/ general\\ SFT datasets}}}] (2) at (7.0,0){$s_2$};
  \node[box 3, label={[xshift=0cm]below:{\parbox{3cm}{\scriptsize Fine-tuned w/ expert-written\\ Judicial Examination datasets (JE-\textsc{Expert})}}}] (3) at (10.5,0){$s_3$};
  \node[box 3, label={[xshift=0cm]below:{\parbox{2.4cm}{\scriptsize Fine-tuned w/ legal consultation datasets}}}] (4) at (14.0,0){$s_4$};

  \node[box 2, label={[xshift=0cm]below:{\parbox{3.3cm}{\scriptsize Fine-tuned w/ ChatGPT-distilled JE datasets, \underline{not} given golden answers (JE-Q2EA)}}}] (5) at (9.75,-1.9){$s_5$};
  \node[box 2, label={[xshift=0cm]below:{\parbox{2.4cm}{\scriptsize Fine-tuned w/ legal consultation datsets}}}] (6) at (13.25,-1.9){$s_6$};

  \node[box 2, label={[xshift=0cm]below:{\parbox{3.3cm}{\scriptsize Fine-tuned w/ ChatGPT-distilled JE datasets, given golden answers\\ (JE-QA2E)}}}] (7) at (9.0,-3.8){$s_7$};
  \node[box 2, label={[xshift=0cm]below:{\parbox{2.4cm}{\scriptsize Fine-tuned w/ legal consultation datssets}}}] (8) at (12.5,-3.8){$s_8$};

  \draw[arrow] (0.east) to (1.west);
  \draw[arrow] (1.east) to (2.west);
  \draw[arrow] (2.east) to (3.west);
  \draw[arrow] (3.east) to (4.west);

  \draw[arrow, out=340, in=180] (2.east) to (5.west);
  \draw[arrow] (5.east) to (6.west);
  
  \draw[arrow, out=305, in=180] (2.east) to (7.west);
  \draw[arrow] (7.east) to (8.west);
\end{tikzpicture}
\caption{The training process of Lawyer LLaMA, where each node $s_i$ represents the $i$-th training $s$tage.}
\label{fig:training_process}
\end{figure*}
\subsection{National Judicial Examination}
\label{sec:National Judicial Examination}
National Judicial Examination is a closed-book exam that applicants must pass to be certified as a lawyer or a judge.  
It is a comprehensive evaluation of legal knowledge and professional skills for legal practitioners. Thus a model could learn legal skills by practicing the questions from such exams. 

We use JEC-QA~\cite{zhong2019jec} to construct our data. JEC-QA includes 26,365 multiple-choice and multiple-answer questions but it does not contain any explanations for the answers. 
To collect SFT examples of Judicial Examination (JE), we first transfer these multiple-choice questions to the question-answering style. We also heuristically remove the examples whose questions are incorrectly transferred. 
We then tried three methods to collect explanations: \textbf{\textsc{JE-Q2EA}}, \textbf{\textsc{JE-QA2E}} and \textbf{\textsc{JE-Expert}}. 
The first two methods use ChatGPT to generate explanations for each question. 
For \textsc{Q2EA} (Question to Explanation+Answer), we input a question into ChatGPT and ask it to output the explanation and answer. 
We find the answers and explanations of \textsc{JE-Q2EA} are often incorrect, so we tried \textsc{JE-QA2E} (Question+Answer to Explanation), where we input both the question and corresponding answer into ChatGPT and ask it to output the explanation only. 
However, the explanations of JE-QA2E are often illogical and hallucinated.
To ensure that both explanations and answers are correct, we collect the analysis of the examination questions in the past two years, which are written by experts. 
In total, we collect 42k examples for JE-Q2EA, 6k ones for JE-QA2E, and 850 ones for \textsc{JE-Expert}.
See the collection details in Appendix~\ref{appendix:fakao} and examples of the three methods in Table~\ref{tab:fakao_version_examples}.

\subsection{Legal Consultation}
\label{sec:legal consultation}
In this paper, we focus on questions about \textbf{marriage} and we use this specific topic as an example to show the effectiveness of our framework. 
We sample seed questions related to marriage from an open-source dataset of legal consultation~\cite{hualv} and ask ChatGPT to act as a lawyer and generate responses. 
We observe that ChatGPT tends to provide legal articles that are outdated or do not exist at all when generating responses, a phenomenon known as hallucination~\cite{Hallucination}.
To improve the faithfulness of generated responses, we adopt a legal article retrieval component, which is described in Section~\ref{sec:Retrieving_Relevant_Legal_Articles}, to search related legal articles for a given question. The top 3 retrieved legal articles are appended to the input prompt, helping ChatGPT to generate more faithful responses.
In total, we collect about 16,000 responses.
See collection details in Appendix~\ref{appendix:legal_consultation} and examples in Table~\ref{tab:article_multi_turn_en}.

\section{Training Process}

\begin{figure*}[t]
\centering
\includegraphics[scale=0.2]{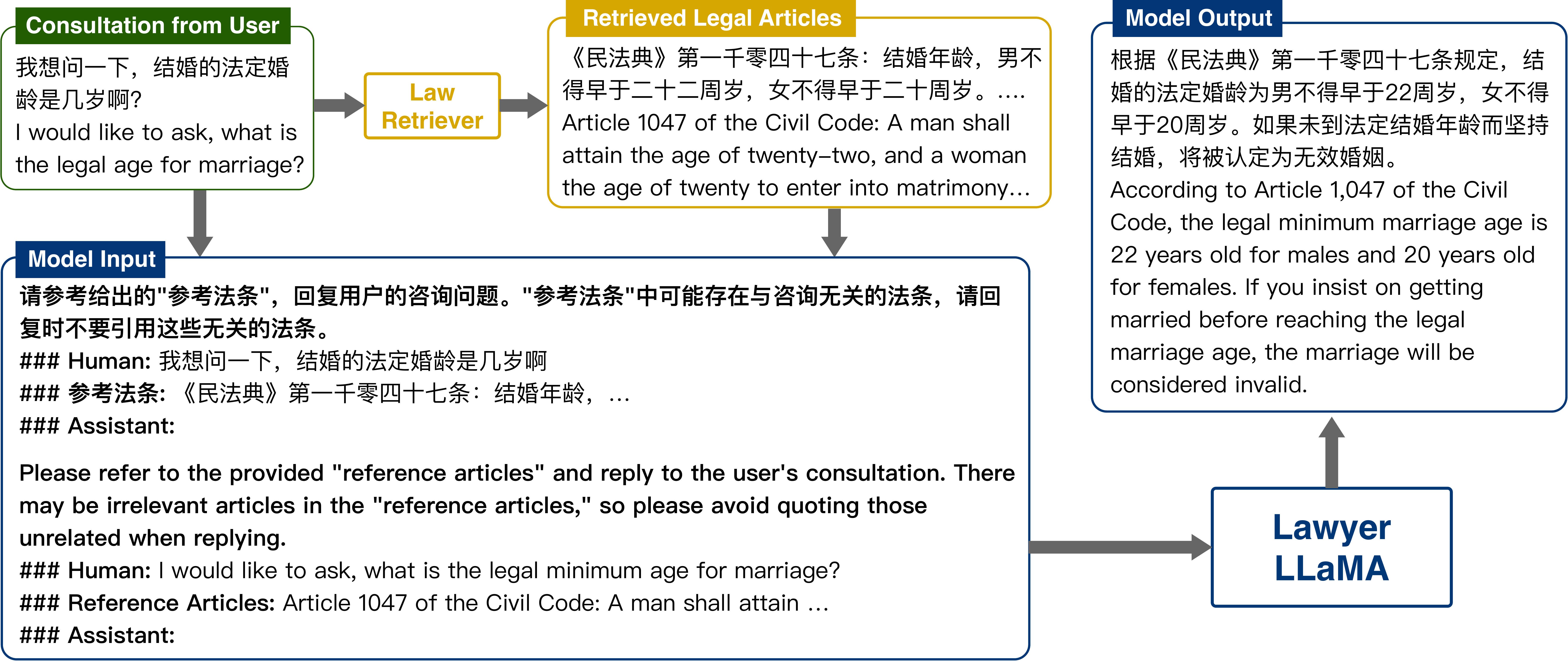}
\caption{An illustration of the inference process of Lawyer LLaMA.}
\label{fig:inference}
\end{figure*}

Figure~\ref{fig:training_process} illustrates how we fine-tune the open-source model LLaMA step by step to adapt it for the Chinese legal domain.

Because few legal domain corpora are used for training Chinese-LLaMA, our first task is to inject legal knowledge into it ($s_1$ in Figure~\ref{fig:training_process}). 
Then, to teach our model to follow human instructions, we train it with general-domain supervised-finetuning datasets ($s_2$ in Figure~\ref{fig:training_process}).
Afterward, we teach the model to reason with the legal knowledge to solve practical problems, by training it on Judicial Examination questions with different versions of CoT ($s_3$, $s_5$ and $s_7$ in Figrue~\ref{fig:training_process}).
Considering that legal consultation for the general public is a typical scenario for using legal models, we 
further tune the model with legal consultation examples ($s_4$, $s_6$ and $s_8$ in Figrue~\ref{fig:training_process}). 
In this stage, to further improve the reliability of the model's responses, we introduce a legal article retrieval module, enabling the model to generate responses with evidence.

\subsection{Injecting Legal Knowledge }
To make up for the deficiency of legal knowledge in the Chinese-LLaMA model, we further pre-train it with Chinese legal Corpus in Section~\ref{sec:zh_legal_corpus}.
This corpus contains a variety of  Chinese texts in the legal domain for further continual training, including legal articles, judicial interpretations, and legal documents. We hope that the vast legal knowledge in the corpus could be injected into Chinese-LLaMA through continual pre-training.
We also add general-domain texts to prevent our model from overfitting the legal corpus.

\subsection{Learning Reasoning Skills}
Solving practical problems requires the model's reasoning skills in the legal domain. 
To this end, we select supervised data from downstream tasks and tune our model with instructions.

We first train our model with about 50,000 instances from Alpaca-GPT-4~\cite{peng2023instruction}, to teach it the general instruction-following ability. It should be noticed that we only use data in Chinese.
Afterward, JE questions with different versions of CoT explanations in Section~\ref{sec:National Judicial Examination} are used for instruction tuning. The JE questions cover a wide range of complex legal questions including distinguishing terminologies and analyzing complicated legal cases. 
We expect that these questions can help our model learn to reason with the legal knowledge learned in continual pre-training. 
Furthermore, we select legal consultation as a real-world scenario for applying our model and train our model with ChatGPT-generated legal consultation responses in Section~\ref{sec:legal consultation}.

\subsection{Retrieving Relevant Legal Articles}
\label{sec:Retrieving_Relevant_Legal_Articles}
As China is a country adopting civil law, legal articles are necessary whenever a judge makes a judgment or a lawyer gives advice. 
However, our preliminary experiment shows that even if the model repeatedly learns the articles during its continual training phase, it cannot use them correctly when generating a response. 
It might refer to an irrelevant article or replace a terminology with a semantically similar word that actually has a drastically different meaning in the legal domain. 
To make the model produce more reliable responses, we incorporate it with a legal article retrieval module.

To train the retrieval module, we first collect user questions for legal consultation and ask legal professionals to annotate up to 3 necessary legal articles to answer each question. 
We then train a legal article retrieval model based on RoBERTa~\cite{liu-etal-2019-roberta}.
It could achieve 0.543 Macro-Recall@1 and 0.807 Macro-Recall@3 on the held-out test set.
In fact, legal article classification is not a new task. Previous works~\cite{zhong-etal-2020-element,zhong-etal-2018-legal} predict the necessary legal articles for judgment based on the fact descriptions in judicial documents. 
We do not directly adopt such classifiers because there is a huge gap between the fact descriptions in judicial documents and real-world user queries. 
The former are formal and well-written while the latter 
The fact descriptions are formal and well-written, are colloquially posed and contain only limited, incomplete information.

When we directly concatenate the retrieved legal articles and the user's question as new input, the model tends to quote all the provided legal articles in its response, without distinguishing whether they are truly relevant to the current scenario. 
Since we don't have a perfect legal retrieval module,  this behavior of the model will introduce more noise.
To address this problem, we add irrelevant articles to the context during training, forcing the model to ignore the distracting information. Furthermore, in the prompt during inference, we inform the model that there might be irrelevant information in the provided legal articles: \textit{There may be irrelevant articles in the reference articles, so please avoid quoting those unrelated ones when replying.}

\section{Automatic Evaluation}
In this section, we use automatic evaluation to investigate whether our framework could help the LLM learn legal knowledge and professional skills. We will further explore the influence of different training steps by human evaluation in Section~\ref{sec:human_evaluation}. 
Another under-explored question is whether injecting domain-specific knowledge will affect the ability of LLMs in general tasks, such as Natural Language Inference (NLI) and Question Answering (QA). To figure it out, we also  evaluate our model on several general benchmarks in both Chinese and English.

Following \citet{eval-harness}, we convert all the tasks into the multi-choice format and select the choice with the lowest perplexity as the model's prediction. The zero-shot performance of our models in different stages is shown in Table~\ref{table:zero_shot_results}.

\begin{table}[t]
\vskip 0.1in
\centering
\tabcolsep=0.3em
\begin{small}
\begin{tabular}{c cc cc cc c}
\toprule
\multirow{2}*{$\bm{s_i}$} & \multicolumn{2}{c}{\textbf{Legal}} & \multicolumn{5}{c}{\textbf{General}}\\ 
\cmidrule(lr){2-3}\cmidrule(lr){4-8}
 & \textbf{CP} & \textbf{JE-M} & \textbf{C3} & \textbf{CMNLI} & \textbf{SciQ} & \textbf{PIQA} & \textbf{Avg.}\\
\midrule
$s_0$ & 18.89 & 49.73 & 49.40 & 31.09 & 89.60 & 76.50 & 61.65\\
$s_1$ & 73.56 & 53.52 & 56.34 & 32.93 & 83.30 & 77.91 & 62.62\\
$s_2$ & 62.22 & 51.61 & 57.81 & 39.20 & 79.80 & 77.75 & 63.64\\
$s_3$ & 78.44 & 57.80 & 55.61 & 40.88 & 82.80 & 77.53 & 64.21\\
$s_4$ & 79.33 & 59.14 & 56.37 & 41.58 & 81.10 & 77.42 & 64.12\\

\bottomrule
\end{tabular}
\end{small}
\caption{\label{table:zero_shot_results}
Model performance of various NLP tasks at different stages, where the details of each $s$tage $s_i$ can refer to Figure \ref{fig:training_process}. The metric is accuracy(\%).
}
\vskip -0.15in
\end{table}

\subsection{Assessing the Ability in Legal Domain}
We use two tasks to evaluate models' ability in the legal domain, Charge Prediction~(CP) and 
National Judicial Examination about Marriage (JE-M). 

Charge prediction is a widely used task to assess whether a model understands legal text~\cite{luobingfeng,confusingcharges}, which aims to predict the charge for a case given its fact description. We choose 9 charges which are hard to distinguish in practice~\cite{Ouyang:99} and randomly sample 100 examples for each charge from CAIL2018~\cite{Xiao2018CAIL2018AL}. See more details in Appendix \ref{sec:legal_cla_details}.

National Judicial Examination is a comprehensive evaluation of legal knowledge and professional skills~\cite{zhong2019jec}. In order to further explore the model's capacity in Marriage, we heuristically select all the multi-choice questions about marriage. 
Considering perplexity-based evaluation can only predict one answer for every question and many cases in National Judicial Examination have multi correct answers, we convert each multiple-choice question to four true-or-false questions by combing the question with every option. In this way, we collect 1,116 examples to form JE-M and it is guaranteed that JE-M does not overlap with the data collected in Section~\ref{sec:National Judicial Examination}.

From the results of Table~\ref{table:zero_shot_results}, we can find that all the efforts in incorporating legal knowledge bring significant improvements on our two benchmarks, where training on general instruction data will slightly diminish the performance of the model. The greatest improvement comes from continual pretraining on the legal domain, where $s_1$ outperforms $s_0$ by 54.67\% and 3.80\% on CP and JE-M respectively. This phenomenon aligns with the findings of previous work, i.e., the model primarily learns knowledge during the pretraining stage~\cite{Hu2022ASO}. As for fine-tuning on domain-specific tasks, it can be seen that $s_3$ and $s_4$ achieve further improvements, implying finetuning on the Judicial Examination dataset and legal consultation dataset might help the model make better use of the knowledge learned in the continual pretraining stage. We also notice that after training on general instruction data, Alpaca-GPT-4, $s_2$ performs slightly worse than $s_1$.  This might be because the methods for solving general problems are not directly applicable to the legal domain.

\subsection{Evaluation on General Tasks}
\label{sec:general_evaluation}
To explore the impact of injecting domain knowledge on the model's ability to solve general tasks, we conduct evaluation on four datasets from general domains. Two of them are Chinese datasets, which are C3~\cite{sun-etal-2020-investigating}, a dialogue Question Answering dataset, and CMNLI~\cite{xu-etal-2020-clue}, a natural language inference dataset. The rest two datasets 
SciQ~\cite{Welbl2017CrowdsourcingMC} and PIQA~\cite{piqa-Bisk2020}, 
are both Question Answering datasets in English. The former is about science and the latter is about commonsense. 

It is exciting to observe that injecting domain knowledge seems not to hurt the model's performance on general tasks.  After pre-training in the legal domain, $s_1$ achieves an average improvement of 0.97\%over $s_0$ in general tasks. Specifically, $s_1$ outperforms $s_0$ in three out of four tasks, with a slight decrease only in the English dataset SciQ. This might be because, in our pre-training corpus, there is only a small amount of English texts, which causes the  decline in the performance in English.

Another interesting finding is that finetuning on legal tasks could enhance the model's reasoning ability. As illustrated in Table~\ref{table:zero_shot_results}, compared with $s_2$, training on the National Judicial Examination makes $s_3$ achieve 1.68\% improvements on CMNLI. And with the help of the legal consultation dataset, $s_4$ further increase the accuracy of CMNLI from 40.88\% to 41.58\%. We believe that, during the learning process of legal tasks, the model has acquired the ability to associate abstract concepts with specific scenarios. This capability has led to improvement in the performance on the natural language inference task.

\section{Human Evaluation}
\label{sec:human_evaluation}
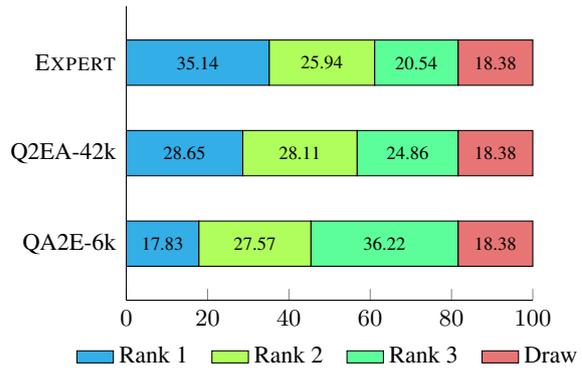
\begin{figure}
\centering
\begin{tikzpicture}
\begin{axis}[
    xbar stacked,
    legend style={
    legend columns=4,
        at={(xticklabel cs:0.5)},
        anchor=north,
        draw=none
    },
    ytick=data,
    axis y line*=none,
    axis x line*=bottom,
    tick label style={font=\footnotesize},
    legend style={font=\footnotesize},
    legend image post style={scale=0.8},
    label style={font=\footnotesize},
    xtick={0,20,40,60,80,100},
    width=0.9\columnwidth,
    bar width=6mm,
    xlabel={Proportion (\%)},
    yticklabels={QA2E-6k, Q2EA-42k, \textsc{Expert}},
    xmin=0,
    xmax=100,
    area legend,
    y=12mm,
    enlarge y limits={abs=0.625},
]
\addplot[black,fill=cmp1!80!cyan!80] coordinates
{(17.83,0) (28.65,1) (35.14,2)};
\addplot[black,fill=gmp1!80!cyan!80] coordinates
{(27.57,0) (28.11,1) (25.94,2)};
\addplot[black,fill=ump1!80!cyan!80] coordinates
{(36.22,0) (24.86,1) (20.54,2)};
\addplot[black,fill=xmp1!54] coordinates
{(18.38,0) (18.38,1) (18.38,2)};
\legend{Rank 1\ \ \ \ , Rank 2\ \ \ \ , Rank 3\ \ \ \ , Draw}
\coordinate (A1) at (8.92,0);
\coordinate (A2) at (31.62,0);
\coordinate (A3) at (63.51,0);
\coordinate (Ad) at (90.81,0);
\coordinate (B1) at (14.33,12mm);
\coordinate (B2) at (42.71,12mm);
\coordinate (B3) at (69.19,12mm);
\coordinate (Bd) at (90.81,12mm);
\coordinate (C1) at (17.57,24mm);
\coordinate (C2) at (48.11,24mm);
\coordinate (C3) at (71.35,24mm);
\coordinate (Cd) at (90.81,24mm);
\end{axis} 
\node at (A1) {\scriptsize{17.83}};
\node at (A2) {\scriptsize{27.57}};
\node at (A3) {\scriptsize{36.22}};
\node at (Ad) {\scriptsize{18.38}};
\node at (B1) {\scriptsize{28.65}};
\node at (B2) {\scriptsize{28.11}};
\node at (B3) {\scriptsize{24.86}};
\node at (Bd) {\scriptsize{18.38}};
\node at (C1) {\scriptsize{35.14}};
\node at (C2) {\scriptsize{25.94}};
\node at (C3) {\scriptsize{20.54}};
\node at (Cd) {\scriptsize{18.38}};
\end{tikzpicture}
\caption{Ranking results of the responses outputted by the models trained with three different versions of JE explanations.}
\label{fig:rank_fako}
\vskip -0.1in
\end{figure}

We conduct three groups of human evaluation experiments to answer the following research questions: \textbf{(1) What kind of SFT data is suitable for domain adaption?
(2) Can retrieval alleviate hallucinations and provide more reliable responses? (3) How to filter out irrelevant information when provided with external knowledge?}

We select 37 real-world consultation questions related to marriage, which cover a wide range of topics, such as divorce, inheritance, and division of property. See the full list in Table~\ref{tab:eval_questions1} and Table~\ref{tab:eval_questions2}.
We recruit 3 annotators with legal backgrounds to rank the output of different models. They can use online search engines or reference books during the evaluation.

\subsection{ What kind of SFT data is suitable?}
\label{sec:sft_quality}
Generating SFT data and distilling knowledge from more powerful LLMs, such as ChatGPT, becomes a new trend in recent months~\cite{wang2022self,yunxiang2023chatdoctor,peng2023instruction}. Some studies also find that the ChatGPT-generated data is more diverse and useful than human-written ones for tasks in the general domain~\cite{alpaca_eval}. Considering that certain domains require a lot of specialized knowledge, which may not be well-captured by general LLMs like ChatGPT, we are curious about what kind of SFT data would be more suitable in the domain adaptation.

To investigate how the quality of SFT datasets affects the model outputs, we ask annotators to rank the outputs from the models trained with three versions of JE explanations, \textsc{JE-Q2EQ}, \textsc{JE-QA2E} and \textsc{JE-Expert}, which are described in detail in Section~\ref{sec:National Judicial Examination}.
If the annotators think that there is no clear difference in the quality of the three outputs, they are required to annotate it as a \textit{draw}. Table~\ref{tab:human_eval_examples_different_je_data} shows an example of outputs from different models to the same question.

The evaluation results are shown in Figure~\ref{fig:rank_fako}. The model trained on only 850 \textsc{JE-Expert} instances achieves the best overall performance, outperforming models trained on 42k ChatGPT-distilled instances, \textsc{JE-Q2EA}. 
It proves that, for learning professional skills, hundreds of expert-written high-quality instructions are more valuable than tens of thousands of ChatGPT-generated ones.

When comparing the two ChatGPT-distilled versions of JE explanation, although JE-QA2E contains fewer errors than JE-Q2EA, the model trained on 42k \textsc{JE-Q2EA} instances outperforms that trained on 6k \textsc{JE-QA2E}. It indicates that, for the automatically generated SFT data, the larger quantity could make up for its lower quality. 

We also wonder whether improving the correctness of ChatGPT-distilled explanations can bring a better model without regard to the influence of dataset scale. Thus, we sample a subset of \textsc{JE-Q2EA} with 6k examples to train another model. 
Then for each query, we collect the responses of models trained with JE-Q2AE-6k or JE-QA2E, and ask annotators to determine which one is better. Among all pairs, annotators believe that \textbf{40.54\%} of the responses generated by JE-QA2E model are better, but merely \textbf{28.38\%} of the responses generated by JE-Q2EA-6k are better, while the rest 31.08\% responses are draw. It indicates with the same scale of SFT dataset, improving the correctness of the explanation can improve model performance.

In this section, we reveal that expert-written examples, whose explanations and conclusions are both correct, are most suitable for teaching the model to learn domain-specific knowlegde and skills. Even we use 50 times more ChatGPT-generated examples, models cannot yet outperform the one trained with expert-written examples. It might be due to the lack of domain-specific knowledge of ChatGPT. 
However, when expert-written examples are not accessible, the scale of ChatGPT-distilled dataset can be very important. And if dataset scales are kept the same, it will still be effective to provide correct answers in prompts when collecting explanations from ChatGPT.

\subsection{Can retrieval alleviate hallucination?}
\label{sec:retrieval_alleviates_hallucinaiton}

\begin{table}[t]
\centering
\begin{footnotesize}
\begin{tabular}{p{0.93\columnwidth}}
\toprule
\textbf{Question:} What is the legal age for marriage in China? \\
\textbf{Correct Response:} According Article 1047 of \textit{the Civil Code},
in China, a man shall attain the age of twenty-two, and a woman the age of twenty to enter into matrimony...\\
\midrule
\textbf{Response with $\bm{\mathcal{H}_1}$:} According \textit{the Civil Code}, \sethlcolor{yellow}\hl{an adult may marriage}, but a minor may not marriage...\\
\midrule
\textbf{Response with $\bm{\mathcal{H}_2}$:} According \sethlcolor{yellow}\hl{Article 32 of \textit{the Marriage and Family Administration Regulations}}, a man should fully attained the age of twenty-two years old to get married, and a woman the age of twenty years old...\\
\bottomrule
\end{tabular}
\caption{\label{tab:hallucination_example_case}
Examples of the two types of hallucinations. \sethlcolor{yellow}\hl{Texts with yellow background} are the nonexistent article-related contents fabricated by model.
}
\end{footnotesize}
\vskip -0.1in
\end{table}

LLMs are likely to generate hallucinated texts~\cite{Hallucination}. It is hard for a person without professional knowledge to distinguish such fluent but unfaithful texts, which might mislead the user and cause unwanted damages, especially in the domain of law. To alleviate such a problem, we augment our Lawyer LLaMA with a retrieval module and in this section, we want to assess whether this module could help the model generate more reliable responses.

As legal articles are the most important evidence in the legal domain, we focus on two types of hallucinations related to legal articles, whose examples are shown in Table~\ref{tab:hallucination_example_case}:

\noindent $\bm{\mathcal{H}_1}$: Whether the model fabricates one or more nonexistent articles.

\noindent $\bm{\mathcal{H}_2}$: If the response mentions an existing article, judge whether it quotes an incorrect title of law or a wrong number of the article.

\begin{table}[t]
\vskip 0.1in
\centering
\begin{small}
\begin{tabular}{c cc}
\toprule
Model & $\mathcal{H}_1$ & $\mathcal{H}_2$\\
\midrule
JE-QA2E & 25.9 & 14.8\\
JE-Q2EA-6k & 64.8 & 60.2\\
\bottomrule
\end{tabular}
\end{small}
\caption{\label{table:hallu_prop}
The proportion~(\%) of responses that contain the two types of hallucinations, among all responses generated by each model.
}
\vskip -0.15in
\end{table}

We use $s_3$, which is trained on JE-\textsc{Expert}, as our base model. We do not use models trained on legal consultation, like $s_5$, because the responses in legal consultation contain possible legal articles and might cause data leakage. For each query, we design two inputs for the model. One is just the query without any other external knowledge and we denote the responses to it as  $\mathbf{r^0}$. 
As for the other, we use the retrieval model introduced in Section \ref{sec:Retrieving_Relevant_Legal_Articles} to provide three corresponding articles and supplement the original input with them. The responses to this type of input are denoted as $\mathbf{r^1}$.

Table~\ref{table:hallu_prop} shows the frequency of these two types of hallucinations that occurs in $\mathbf{r^0}$ and $\mathbf{r^1}$.
For all responses in $\mathbf{r^1}$ which mention articles, \textbf{25.9\%} of them have nonexistent articles, and \textbf{14.8\%} of them quote incorrect titles or article numbers. However, for the responses in $\mathbf{r^0}$ mentioning articles, \textbf{64.8\%} of them use nonexistent articles, and \textbf{60.2\%} of them quote wrong titles or numbers. The obvious gap shows the importance of the retrieval module in reducing hallucinations about legal articles. Although we cannot eliminate hallucinations only by providing corresponding articles, it is a simple but effective method to make models generate reliable responses which comply with the law.

\subsection{How to filter out irrelevant information?}
\label{sec:effect_of_denoising}
When provided with external knowledge as extra input, we find the models are likely to use all of them, without discriminating whether the provided information is really needed. However, as discussed in Section \ref{sec:Retrieving_Relevant_Legal_Articles}, our retrieval model is not perfect, with 0.807 Macro-Recall@3, which means that many unnecessary articles will also be retrieved as extra input.

To help the model make better use of retrieved articles and filter out unnecessary ones, we add extra irrelevant articles to the model input when fine-tuning with legal consultation examples. We hope the model can learn to discriminate the corresponding articles and irrelevant ones. We take the model at $s_4$ as our baseline. As a comparison, we also fine-tune a model from $s_3$ with the legal consultation examples which only include golden articles. We denote this model as $s_4^{\text{golden}}$.

We then ask human annotators to judge whether the model utilizes irrelevant articles and responses redundantly.
Table~\ref{tab:redundant_example_case} gives an example of a redundant response. 
The query is not relevant to how to adopt a child. However, since Article 1103 of the Civil Code is extracted by the retrieval module, the model might mistakenly use it to generate the response, making the response off-topic.
In our experiment, the percentage of responses  generated by $s_4^{\text{golden}}$ being annotated as redundant is 26.1\%, which is nearly double that of $s_4$
The huge gap illustrates that adding irrelevant evidence during training is important and effective to filter out unnecessary information when a domain-specific model is augmented with a retrieval module.

\begin{table}[t]
\centering
\begin{footnotesize}
\begin{tabular}{p{0.93\columnwidth}}
\toprule
\textbf{Input:} Reference articles: (1) \uline{Article 1072 of the Civil Code}: ...; (2) \uline{Article 26 of the Civil Code}: ...; (3) \uline{Article 1103 of the Civil Code}: ... Please answer following question based on the given articles: \textit{Do I have an obligation to provide financial support for my stepdaughter?} \\
\midrule
\textbf{Response without unnecessary articles:} According \uline{Article 1072 and Article 26 of \textit{the Civil Code}}, the stepparents bear the duty to raise their stepchildren. Therefore, you should provide financial support to your stepdaughter.\\
\midrule
\textbf{Redundant response:} According to \uline{Article 1072 of \textit{the Civil Code}}, the duties and rights between a stepparent and a stepchild are the same as that between parents and children. According to \uline{Article 26 of \textit{the Civil Code}}, parents bear the duty to raise their children. \sethlcolor{yellow}\hl{According to Article 1103 of \textit{the Civil Code}, a stepparent may, with the consent of the stepchild's biological parents, adopt the stepchild.} I wonder whether you have obtained the consent of your stepdaughter's original parents. If you have, you should provide financial support. Otherwise, you are not obliged to raise your stepdaughter.\\
\bottomrule
\end{tabular}
\caption{\label{tab:redundant_example_case}
Examples of responses with or without unnecessary articles. \sethlcolor{yellow}\hl{Texts with yellow background} are unnecessary articles mentioned in the responses. The model uses it only because it is provided as an extra input.
}
\end{footnotesize}

\end{table}

\section{Related Works}
Previous efforts have tried many methods to take advantage of pre-trained language models to solve legal tasks. LEGAL-BERT~\cite{DBLP:journals/corr/abs-2010-02559} explored applying BERT models to downstream legal tasks using EU and UK legal corpus. Lawformer~\cite{lawformer} is a longformer-based pre-trained language model, specifically further pre-trained on Chinese judicial documents. CaseLawBERT~\cite{caselawbert}
is pre-trained on a corpus of US case law documents and contracts. 
These models exhibit limited capabilities in understanding and applying legal knowledge, creating a gap between them and large language models.

When it comes to the era of Large Language Models, there is no effort to build a legal domain LLM in the literature.
There have been a series of works adapting large language models to other specialized domains.
Med-PaLM~\cite{Med-PaLM} and Med-PaLM 2~\cite{Med-PaLM-2} are two instruction-tuned PaLM~\cite{chowdhery2022palm} models for the medical domain. ChatDoctor~\cite{yunxiang2023chatdoctor} is another model in the medical domain fine-tuned on LLaMA, augmented with a disease database for more reliable responses.
In terms of the financial technology domain, BloombergGPT~\cite{wu2023bloomberggpt} is trained from scratch on a mixture of financial documents and general-domain corpora. As far as we know, we are the first to explore how to build a legal domain-specific LLM.

\section{Conclusion}
This paper proposes a method of applying LLMs to specific domains. First, the model needs to be pre-trained on a large amount of domain-related text to help the model learn the necessary knowledge. Then, we need to design domain-specific tasks and use supervised finetuning to teach the model how to solve practical problems. Finally, we find that a retrieval model providing external evidence would be helpful to alleviate the hallucination problem and improve the reliability of the responses.

\section*{Limitations}
First, most of our work only focuses on the \textit{marriage}-related problems in the legal domain because of the high cost of collecting resources, including annotating related legal articles for consultation, constructing suitable evaluation benchmarks, and evaluating model outputs manually. Yet we believe that our framework can be applied to other topics in the legal domain, such as civil law, criminal law, administrative law, and procedural law. We leave this as future work.

Second, we did not do much quantitative analysis of how the size of human-annotated or automatically-generated SFT data is related to the model performance.
We collected only 850 human-annotated SFT instances (\textsc{JE-Expert}) due to the high cost of human annotation. It is unclear how the model performance changes when we scale up the human-annotated SFT data. 
Besides, as a result of limited computational budgets, we only conducted experiments with 6k and 42k automatically-generated data (JE-Q2EA and JE-QA2E). It would be interesting to quantitatively investigate how much automatically-generated data are sufficient for training a model that can outperform the model trained on hundreds of human-annotated data.

\section*{Ethics Statement}
The purpose of this paper is to explore how large language models can be applied to specific domains, with the legal domain chosen as an example. However, it should be noted that the data we have collected has not been rigorously vetted and may contain erroneous content, which should be used with caution. More importantly, the output of our trained model is not professional legal advice and may include errors. If one needs legal assistance, please seek help from professionals.

\bibliography{anthology,custom}

\begin{thebibliography}{43}
\expandafter\ifx\csname natexlab\endcsname\relax\def\natexlab#1{#1}\fi

\bibitem[{Beeching et~al.(2023)Beeching, Han, Lambert, Rajani, Sanseviero, Tunstall, and Wolf}]{open-llm-leaderboard}
Edward Beeching, Sheon Han, Nathan Lambert, Nazneen Rajani, Omar Sanseviero, Lewis Tunstall, and Thomas Wolf. 2023.
\newblock Open {LLM} {L}eaderboard.
\newblock \url{https://huggingface.co/spaces/HuggingFaceH4/open_llm_leaderboard}.

\bibitem[{Bisk et~al.(2020)Bisk, Zellers, Bras, Gao, and Choi}]{piqa-Bisk2020}
Yonatan Bisk, Rowan Zellers, Ronan~Le Bras, Jianfeng Gao, and Yejin Choi. 2020.
\newblock Piqa: Reasoning about physical commonsense in natural language.
\newblock In \emph{Thirty-Fourth AAAI Conference on Artificial Intelligence}.

\bibitem[{Chalkidis et~al.(2020)Chalkidis, Fergadiotis, Malakasiotis, Aletras, and Androutsopoulos}]{DBLP:journals/corr/abs-2010-02559}
Ilias Chalkidis, Manos Fergadiotis, Prodromos Malakasiotis, Nikolaos Aletras, and Ion Androutsopoulos. 2020.
\newblock \href {http://arxiv.org/abs/2010.02559} {{LEGAL-BERT:} the muppets straight out of law school}.
\newblock \emph{CoRR}, abs/2010.02559.

\bibitem[{Chen et~al.(2023)Chen, Takamura, Kobayashi, and Miyao}]{chen-etal-2023-improving}
Chung-Chi Chen, Hiroya Takamura, Ichiro Kobayashi, and Yusuke Miyao. 2023.
\newblock \href {https://aclanthology.org/2023.findings-eacl.4} {Improving numeracy by input reframing and quantitative pre-finetuning task}.
\newblock In \emph{Findings of the Association for Computational Linguistics: EACL 2023}, pages 69--77, Dubrovnik, Croatia. Association for Computational Linguistics.

\bibitem[{Chen(2018)}]{hualv}
Feida Chen. 2018.
\newblock \href {https://doi.org/10.18170/DVN/OLO4G8} {{The legal consultation data and corpus of the thesis from China law network.Replication Data for: Design and research of legal consultation text classification system.}}

\bibitem[{Chen et~al.(2020)Chen, Hou, Cui, Che, Liu, and Yu}]{chen-etal-2020-recall}
Sanyuan Chen, Yutai Hou, Yiming Cui, Wanxiang Che, Ting Liu, and Xiangzhan Yu. 2020.
\newblock \href {https://doi.org/10.18653/v1/2020.emnlp-main.634} {Recall and learn: Fine-tuning deep pretrained language models with less forgetting}.
\newblock In \emph{Proceedings of the 2020 Conference on Empirical Methods in Natural Language Processing (EMNLP)}, pages 7870--7881, Online. Association for Computational Linguistics.

\bibitem[{Chiang et~al.(2023)Chiang, Li, Lin, Sheng, Wu, Zhang, Zheng, Zhuang, Zhuang, Gonzalez, Stoica, and Xing}]{vicuna2023}
Wei-Lin Chiang, Zhuohan Li, Zi~Lin, Ying Sheng, Zhanghao Wu, Hao Zhang, Lianmin Zheng, Siyuan Zhuang, Yonghao Zhuang, Joseph~E. Gonzalez, Ion Stoica, and Eric~P. Xing. 2023.
\newblock \href {https://lmsys.org/blog/2023-03-30-vicuna/} {Vicuna: An open-source chatbot impressing gpt-4 with 90\%* chatgpt quality}.

\bibitem[{Chowdhery et~al.(2022)Chowdhery, Narang, Devlin, Bosma, Mishra, Roberts, Barham, Chung, Sutton, Gehrmann et~al.}]{chowdhery2022palm}
Aakanksha Chowdhery, Sharan Narang, Jacob Devlin, Maarten Bosma, Gaurav Mishra, Adam Roberts, Paul Barham, Hyung~Won Chung, Charles Sutton, Sebastian Gehrmann, et~al. 2022.
\newblock Palm: Scaling language modeling with pathways.
\newblock \emph{arXiv preprint arXiv:2204.02311}.

\bibitem[{Cui et~al.(2023)Cui, Yang, and Yao}]{chinese-llama-alpaca}
Yiming Cui, Ziqing Yang, and Xin Yao. 2023.
\newblock \href {https://arxiv.org/abs/2304.08177} {Efficient and effective text encoding for chinese llama and alpaca}.
\newblock \emph{arXiv preprint arXiv:2304.08177}.

\bibitem[{d'Autume et~al.(2019)d'Autume, Ruder, Kong, and Yogatama}]{d2019episodic}
Cyprien de~Masson d'Autume, Sebastian Ruder, Lingpeng Kong, and Dani Yogatama. 2019.
\newblock Episodic memory in lifelong language learning.
\newblock In \emph{Proceedings of the 33rd International Conference on Neural Information Processing Systems}, pages 13132--13141.

\bibitem[{Gao et~al.(2021)Gao, Tow, Biderman, Black, DiPofi, Foster, Golding, Hsu, McDonell, Muennighoff, Phang, Reynolds, Tang, Thite, Wang, Wang, and Zou}]{eval-harness}
Leo Gao, Jonathan Tow, Stella Biderman, Sid Black, Anthony DiPofi, Charles Foster, Laurence Golding, Jeffrey Hsu, Kyle McDonell, Niklas Muennighoff, Jason Phang, Laria Reynolds, Eric Tang, Anish Thite, Ben Wang, Kevin Wang, and Andy Zou. 2021.
\newblock \href {https://doi.org/10.5281/zenodo.5371628} {A framework for few-shot language model evaluation}.

\bibitem[{Hu et~al.(2022)Hu, Liu, Zhao, Hou, Nie, and Li}]{Hu2022ASO}
Linmei Hu, Zeyi Liu, Ziwang Zhao, Lei Hou, Liqiang Nie, and Juanzi Li. 2022.
\newblock A survey of knowledge-enhanced pre-trained language models.
\newblock \emph{ArXiv}, abs/2211.05994.

\bibitem[{Ji et~al.(2023)Ji, Lee, Frieske, Yu, Su, Xu, Ishii, Bang, Madotto, and Fung}]{Hallucination}
Ziwei Ji, Nayeon Lee, Rita Frieske, Tiezheng Yu, Dan Su, Yan Xu, Etsuko Ishii, Ye~Jin Bang, Andrea Madotto, and Pascale Fung. 2023.
\newblock \href {https://doi.org/10.1145/3571730} {Survey of hallucination in natural language generation}.
\newblock \emph{ACM Comput. Surv.}, 55(12).

\bibitem[{Lee et~al.(2019)Lee, Yoon, Kim, Kim, Kim, So, and Kang}]{biobert}
Jinhyuk Lee, Wonjin Yoon, Sungdong Kim, Donghyeon Kim, Sunkyu Kim, Chan~Ho So, and Jaewoo Kang. 2019.
\newblock \href {https://doi.org/10.1093/bioinformatics/btz682} {{BioBERT: a pre-trained biomedical language representation model for biomedical text mining}}.
\newblock \emph{Bioinformatics}, 36(4):1234--1240.

\bibitem[{Li et~al.(2023{\natexlab{a}})Li, Zhang, Dubois, Taori, Gulrajani, Guestrin, Liang, and Hashimoto}]{alpaca_eval}
Xuechen Li, Tianyi Zhang, Yann Dubois, Rohan Taori, Ishaan Gulrajani, Carlos Guestrin, Percy Liang, and Tatsunori~B. Hashimoto. 2023{\natexlab{a}}.
\newblock Alpacaeval: An automatic evaluator of instruction-following models.
\newblock \url{https://github.com/tatsu-lab/alpaca_eval}.

\bibitem[{Li et~al.(2023{\natexlab{b}})Li, Li, Zhang, Dan, and Zhang}]{yunxiang2023chatdoctor}
Yunxiang Li, Zihan Li, Kai Zhang, Ruilong Dan, and You Zhang. 2023{\natexlab{b}}.
\newblock \href {http://arxiv.org/abs/2303.14070} {Chatdoctor: A medical chat model fine-tuned on llama model using medical domain knowledge}.

\bibitem[{Liu et~al.(2019)Liu, Ott, Goyal, Du, Joshi, Chen, Levy, Lewis, Zettlemoyer, and Stoyanov}]{liu-etal-2019-roberta}
Yinhan Liu, Myle Ott, Naman Goyal, Jingfei Du, Mandar Joshi, Danqi Chen, Omer Levy, Mike Lewis, Luke Zettlemoyer, and Veselin Stoyanov. 2019.
\newblock \href {http://arxiv.org/abs/1907.11692} {Ro{BERT}a: {A} robustly optimized {BERT} pretraining approach}.
\newblock \emph{CoRR}, abs/1907.11692.

\bibitem[{Luo et~al.(2017)Luo, Feng, Xu, Zhang, and Zhao}]{luobingfeng}
Bingfeng Luo, Yansong Feng, Jianbo Xu, Xiang Zhang, and Dongyan Zhao. 2017.
\newblock \href {https://doi.org/10.18653/v1/d17-1289} {Learning to predict charges for criminal cases with legal basis}.
\newblock In \emph{Proceedings of the 2017 Conference on Empirical Methods in Natural Language Processing, {EMNLP} 2017, Copenhagen, Denmark, September 9-11, 2017}, pages 2727--2736. Association for Computational Linguistics.

\bibitem[{Ouyang(1999)}]{Ouyang:99}
Tao Ouyang. 1999.
\newblock \emph{Confusing crimes, noncrime, and boundaries between crimes}, volume~1.

\bibitem[{Peng et~al.(2023)Peng, Li, He, Galley, and Gao}]{peng2023instruction}
Baolin Peng, Chunyuan Li, Pengcheng He, Michel Galley, and Jianfeng Gao. 2023.
\newblock Instruction tuning with gpt-4.
\newblock \emph{arXiv preprint arXiv:2304.03277}.

\bibitem[{Raffel et~al.(2020)Raffel, Shazeer, Roberts, Lee, Narang, Matena, Zhou, Li, and Liu}]{t5-c4}
Colin Raffel, Noam Shazeer, Adam Roberts, Katherine Lee, Sharan Narang, Michael Matena, Yanqi Zhou, Wei Li, and Peter~J. Liu. 2020.
\newblock \href {http://jmlr.org/papers/v21/20-074.html} {Exploring the limits of transfer learning with a unified text-to-text transformer}.
\newblock \emph{Journal of Machine Learning Research}, 21(140):1--67.

\bibitem[{Shu et~al.(2023)Shu, Chen, Liu, Fu, Shareghi, and Collier}]{Visual-Med-Alpaca}
Chang Shu, Baian Chen, Fangyu Liu, Zihao Fu, Ehsan Shareghi, and Nigel Collier. 2023.
\newblock Visual med-alpaca: A parameter-efficient biomedical llm with visual capabilities.
\newblock \url{https://github.com/cambridgeltl/visual-med-alpaca}.

\bibitem[{Singhal et~al.(2022)Singhal, Azizi, Tu, Mahdavi, Wei, Chung, Scales, Tanwani, Cole-Lewis, Pfohl et~al.}]{Med-PaLM}
Karan Singhal, Shekoofeh Azizi, Tao Tu, S~Sara Mahdavi, Jason Wei, Hyung~Won Chung, Nathan Scales, Ajay Tanwani, Heather Cole-Lewis, Stephen Pfohl, et~al. 2022.
\newblock Large language models encode clinical knowledge.
\newblock \emph{arXiv preprint arXiv:2212.13138}.

\bibitem[{Singhal et~al.(2023)Singhal, Tu, Gottweis, Sayres, Wulczyn, Hou, Clark, Pfohl, Cole-Lewis, Neal et~al.}]{Med-PaLM-2}
Karan Singhal, Tao Tu, Juraj Gottweis, Rory Sayres, Ellery Wulczyn, Le~Hou, Kevin Clark, Stephen Pfohl, Heather Cole-Lewis, Darlene Neal, et~al. 2023.
\newblock Towards expert-level medical question answering with large language models.
\newblock \emph{arXiv preprint arXiv:2305.09617}.

\bibitem[{Sun et~al.(2020)Sun, Yu, Yu, and Cardie}]{sun-etal-2020-investigating}
Kai Sun, Dian Yu, Dong Yu, and Claire Cardie. 2020.
\newblock \href {https://doi.org/10.1162/tacl_a_00305} {Investigating prior knowledge for challenging {C}hinese machine reading comprehension}.
\newblock \emph{Transactions of the Association for Computational Linguistics}, 8:141--155.

\bibitem[{Tao et~al.(2023)Tao, Feng, and Zhao}]{tao2023can}
Mingxu Tao, Yansong Feng, and Dongyan Zhao. 2023.
\newblock \href {https://openreview.net/forum?id=UazgYBMS9-W} {Can {BERT} refrain from forgetting on sequential tasks? a probing study}.
\newblock In \emph{The Eleventh International Conference on Learning Representations}.

\bibitem[{Taori et~al.(2023)Taori, Gulrajani, Zhang, Dubois, Li, Guestrin, Liang, and Hashimoto}]{alpaca}
Rohan Taori, Ishaan Gulrajani, Tianyi Zhang, Yann Dubois, Xuechen Li, Carlos Guestrin, Percy Liang, and Tatsunori~B. Hashimoto. 2023.
\newblock Stanford alpaca: An instruction-following llama model.
\newblock \url{https://github.com/tatsu-lab/stanford_alpaca}.

\bibitem[{Wang et~al.(2022)Wang, Kordi, Mishra, Liu, Smith, Khashabi, and Hajishirzi}]{wang2022self}
Yizhong Wang, Yeganeh Kordi, Swaroop Mishra, Alisa Liu, Noah~A Smith, Daniel Khashabi, and Hannaneh Hajishirzi. 2022.
\newblock Self-instruct: Aligning language model with self generated instructions.
\newblock \emph{arXiv preprint arXiv:2212.10560}.

\bibitem[{Welbl et~al.(2017)Welbl, Liu, and Gardner}]{Welbl2017CrowdsourcingMC}
Johannes Welbl, Nelson~F. Liu, and Matt Gardner. 2017.
\newblock Crowdsourcing multiple choice science questions.
\newblock \emph{ArXiv}, abs/1707.06209.

\bibitem[{Wu et~al.(2023)Wu, Irsoy, Lu, Dabravolski, Dredze, Gehrmann, Kambadur, Rosenberg, and Mann}]{wu2023bloomberggpt}
Shijie Wu, Ozan Irsoy, Steven Lu, Vadim Dabravolski, Mark Dredze, Sebastian Gehrmann, Prabhanjan Kambadur, David Rosenberg, and Gideon Mann. 2023.
\newblock Bloomberggpt: A large language model for finance.
\newblock \emph{arXiv preprint arXiv:2303.17564}.

\bibitem[{Xiao et~al.(2021)Xiao, Hu, Liu, Tu, and Sun}]{lawformer}
Chaojun Xiao, Xueyu Hu, Zhiyuan Liu, Cunchao Tu, and Maosong Sun. 2021.
\newblock \href {http://arxiv.org/abs/2105.03887} {Lawformer: {A} pre-trained language model for chinese legal long documents}.
\newblock \emph{CoRR}, abs/2105.03887.

\bibitem[{Xiao et~al.(2018)Xiao, Zhong, Guo, Tu, Liu, Sun, Feng, Han, Hu, Wang, and Xu}]{Xiao2018CAIL2018AL}
Chaojun Xiao, Haoxiang Zhong, Zhipeng Guo, Cunchao Tu, Zhiyuan Liu, Maosong Sun, Yansong Feng, Xianpei Han, Zhen Hu, Heng Wang, and Jianfeng Xu. 2018.
\newblock Cail2018: A large-scale legal dataset for judgment prediction.
\newblock \emph{ArXiv}, abs/1807.02478.

\bibitem[{Xiong et~al.(2023)Xiong, Wang, Zhu, Zhao, Liu, Wang, and Shen}]{xiong2023doctorglm}
Honglin Xiong, Sheng Wang, Yitao Zhu, Zihao Zhao, Yuxiao Liu, Qian Wang, and Dinggang Shen. 2023.
\newblock Doctorglm: Fine-tuning your chinese doctor is not a herculean task.
\newblock \emph{arXiv preprint arXiv:2304.01097}.

\bibitem[{Xu et~al.(2020{\natexlab{a}})Xu, Hu, Zhang, Li, Cao, Li, Xu, Sun, Yu, Yu, Tian, Dong, Liu, Shi, Cui, Li, Zeng, Wang, Xie, Li, Patterson, Tian, Zhang, Zhou, Liu, Zhao, Zhao, Yue, Zhang, Yang, Richardson, and Lan}]{xu-etal-2020-clue}
Liang Xu, Hai Hu, Xuanwei Zhang, Lu~Li, Chenjie Cao, Yudong Li, Yechen Xu, Kai Sun, Dian Yu, Cong Yu, Yin Tian, Qianqian Dong, Weitang Liu, Bo~Shi, Yiming Cui, Junyi Li, Jun Zeng, Rongzhao Wang, Weijian Xie, Yanting Li, Yina Patterson, Zuoyu Tian, Yiwen Zhang, He~Zhou, Shaoweihua Liu, Zhe Zhao, Qipeng Zhao, Cong Yue, Xinrui Zhang, Zhengliang Yang, Kyle Richardson, and Zhenzhong Lan. 2020{\natexlab{a}}.
\newblock \href {https://doi.org/10.18653/v1/2020.coling-main.419} {{CLUE}: A {C}hinese language understanding evaluation benchmark}.
\newblock In \emph{Proceedings of the 28th International Conference on Computational Linguistics}, pages 4762--4772, Barcelona, Spain (Online). International Committee on Computational Linguistics.

\bibitem[{Xu et~al.(2020{\natexlab{b}})Xu, Zhang, and Dong}]{xu2020cluecorpus2020}
Liang Xu, Xuanwei Zhang, and Qianqian Dong. 2020{\natexlab{b}}.
\newblock \href {http://arxiv.org/abs/2003.01355} {Cluecorpus2020: A large-scale chinese corpus for pre-training language model}.

\bibitem[{Xu et~al.(2020{\natexlab{c}})Xu, Wang, Chen, Pan, Wang, and Zhao}]{confusingcharges}
Nuo Xu, Pinghui Wang, Long Chen, Li~Pan, Xiaoyan Wang, and Junzhou Zhao. 2020{\natexlab{c}}.
\newblock \href {https://doi.org/10.18653/v1/2020.acl-main.280} {Distinguish confusing law articles for legal judgment prediction}.
\newblock In \emph{Proceedings of the 58th Annual Meeting of the Association for Computational Linguistics, {ACL} 2020, Online, July 5-10, 2020}, pages 3086--3095. Association for Computational Linguistics.

\bibitem[{Yuan et~al.(2021)Yuan, Zhao, Du, Ding, Liu, Cen, Zou, Yang, and Tang}]{wudao-yuan-et-al}
Sha Yuan, Hanyu Zhao, Zhengxiao Du, Ming Ding, Xiao Liu, Yukuo Cen, Xu~Zou, Zhilin Yang, and Jie Tang. 2021.
\newblock \href {https://doi.org/https://doi.org/10.1016/j.aiopen.2021.06.001} {Wudaocorpora: A super large-scale chinese corpora for pre-training language models}.
\newblock \emph{AI Open}, 2:65--68.

\bibitem[{Zeng et~al.(2023)Zeng, Liu, Du, Wang, Lai, Ding, Yang, Xu, Zheng, Xia, Tam, Ma, Xue, Zhai, Chen, Liu, Zhang, Dong, and Tang}]{zeng2023glm-130b}
Aohan Zeng, Xiao Liu, Zhengxiao Du, Zihan Wang, Hanyu Lai, Ming Ding, Zhuoyi Yang, Yifan Xu, Wendi Zheng, Xiao Xia, Weng~Lam Tam, Zixuan Ma, Yufei Xue, Jidong Zhai, Wenguang Chen, Zhiyuan Liu, Peng Zhang, Yuxiao Dong, and Jie Tang. 2023.
\newblock \href {https://openreview.net/forum?id=-Aw0rrrPUF} {{GLM}-130b: An open bilingual pre-trained model}.
\newblock In \emph{The Eleventh International Conference on Learning Representations (ICLR)}.

\bibitem[{Zheng et~al.(2023)Zheng, Chiang, Sheng, Zhuang, Wu, Zhuang, Lin, Li, Li, Xing et~al.}]{zheng2023judging}
Lianmin Zheng, Wei-Lin Chiang, Ying Sheng, Siyuan Zhuang, Zhanghao Wu, Yonghao Zhuang, Zi~Lin, Zhuohan Li, Dacheng Li, Eric Xing, et~al. 2023.
\newblock Judging llm-as-a-judge with mt-bench and chatbot arena.
\newblock \emph{arXiv preprint arXiv:2306.05685}.

\bibitem[{Zheng et~al.(2021)Zheng, Guha, Anderson, Henderson, and Ho}]{caselawbert}
Lucia Zheng, Neel Guha, Brandon~R. Anderson, Peter Henderson, and Daniel~E. Ho. 2021.
\newblock \href {https://doi.org/10.1145/3462757.3466088} {When does pretraining help?: assessing self-supervised learning for law and the casehold dataset of 53, 000+ legal holdings}.
\newblock In \emph{{ICAIL} '21: Eighteenth International Conference for Artificial Intelligence and Law, S{\~{a}}o Paulo Brazil, June 21 - 25, 2021}, pages 159--168. {ACM}.

\bibitem[{Zhong et~al.(2018)Zhong, Guo, Tu, Xiao, Liu, and Sun}]{zhong-etal-2018-legal}
Haoxi Zhong, Zhipeng Guo, Cunchao Tu, Chaojun Xiao, Zhiyuan Liu, and Maosong Sun. 2018.
\newblock \href {https://doi.org/10.18653/v1/D18-1390} {Legal judgment prediction via topological learning}.
\newblock In \emph{Proceedings of the 2018 Conference on Empirical Methods in Natural Language Processing}, pages 3540--3549, Brussels, Belgium. Association for Computational Linguistics.

\bibitem[{Zhong et~al.(2020{\natexlab{a}})Zhong, Xiao, Tu, Zhang, Liu, and Sun}]{zhong2019jec}
Haoxi Zhong, Chaojun Xiao, Cunchao Tu, Tianyang Zhang, Zhiyuan Liu, and Maosong Sun. 2020{\natexlab{a}}.
\newblock Jec-qa: A legal-domain question answering dataset.
\newblock In \emph{Proceedings of AAAI}.

\bibitem[{Zhong et~al.(2020{\natexlab{b}})Zhong, Zhou, Qu, Long, and Gu}]{zhong-etal-2020-element}
Huilin Zhong, Junsheng Zhou, Weiguang Qu, Yunfei Long, and Yanhui Gu. 2020{\natexlab{b}}.
\newblock \href {https://doi.org/10.18653/v1/2020.emnlp-main.540} {An element-aware multi-representation model for law article prediction}.
\newblock In \emph{Proceedings of the 2020 Conference on Empirical Methods in Natural Language Processing (EMNLP)}, pages 6663--6668, Online. Association for Computational Linguistics.

\end{thebibliography}
\bibliographystyle{acl_natbib}

\clearpage
\appendix

\section{Details of Data Collection}
\subsection{National Judicial Examination}
\label{appendix:fakao}

To enable the model to apply legal knowledge, we construct supervised fine-tuning datasets based on JEC-QA~\cite{zhong2019jec}, a collection of Chinese National Judicial Examination. All examples in JEC-QA are multiple-choice questions, part of which have more than one correct option. We first try to input the whole question with its four options into ChatGPT. We ask ChatGPT first to generate reasoning explanations by chain-of-thought and then to predict answers. There are two main flaws in such a process: (1) ChatGPT usually outputs incorrect predictions; (2) ChatGPT does only give reasoning analysis for part of the options. We hope Lawyer LLaMA can learn how to discriminate similar legal concepts and how to apply legal knowledge to solve case-analysis questions. Thus, the quality of explanations in the SFT dataset can be important, which means it is improper to train the model with incorrect or uninformative explanations.

To achieve the analysis for each option, we concatenate a question with its four options respectively as four separate inputs for ChatGPT. We also find ChatGPT can generate more coherent responses, if we transfer the concatenated question-answer pair to a single fluent query. We illustrate the transferring process below.

\textbf{Question:} \textit{After one party to the contract has paid the other party compensation for breach of the contract, which of the following options should be taken?}

\textbf{Option:} \textit{It is up to the compensating party to decide whether to continue the performance of the contract.} 

We then should transfer this example to the following query: \textit{After one party to the contract has paid the other party compensation for breach of the contract, can the compensating party decide whether to continue the performance of the contract?}

We first utilize regular expressions to transfer the examples in JEC-QA. Among all question-option pairs, 39.9\% of them can be processed by our regular expressions, while the rest are removed. We then input the transferred queries into ChatGPT for explanations and answers by chain-of-thought. In this way, we collect 42k examples, denoted as JE-Q2EA.

We further find the conclusions of around 1/4 JE-Q2EA examples are incorrect, and more than 2/3 of transferred queries are not fluent and unnatural. Therefore, we try to transfer the question-option pair by in-context learning based on ChatGPT. We use the following prompt.

\textbf{Prompt: }\textit{Here is a question from the National Judicial Examination and one of its options. The question texts will begin with "\#\#\# Question:" and the option texts with "\#\#\# Option:". Please concatenate them as a single true-or-false question. A few examples are illustrated below.}

\textit{\# Example 1:}

\textit{\#\#\# Question: \texttt{[Question 1]}}

\textit{\#\#\# Option: \texttt{[Option 1]}}

\textit{Output: \texttt{[A human-annotated result.]}}

\textit{\# Example 2:}

\textit{\#\#\# Question: \texttt{[Question 2]}}

\textit{\#\#\# Option: \texttt{[Option 2]}}

\textit{Output: \texttt{[A human-annotated result.]}}

\textit{\# Example 3:}

\textit{\#\#\# Question: \texttt{[Question 3]}}

\textit{\#\#\# Option: \texttt{[Option 3]}}

\textit{Output: \texttt{[A human-annotated result.]}}

\textit{Now, please concatenate the following question and option as a single fluent and coherent query. The output should not contain phrases like "options", "from following", "as below".}

\textit{\#\#\# Question: \texttt{[Input Question]}}

\textit{\#\#\# Option: \texttt{[Input Option]}}

\textit{Output:}

Almost all of the questions transferred by ChatGPT can keep the same meanings as the original question-option pairs. However, we find there are still many words indicating the queries come from multiple-choice questions, such as the word \textit{the option} in \textit{After one party to the contract has paid the other party compensation for breach of the contract, \textbf{the option} "the compensating party can decide whether to continue performance of the contract". Is it correct?"}

Such examples are unnatural and confusing, since there are no contexts about other options. If ChatGPT is given these examples, it might reply "I cannot find other options here, so I cannot answer which option should be chosen." To guarantee the quality of the dataset, we filter out the transferred queries containing words like "option", "from following", etc. In this way, it remains 6k transferred queries. We then input the queries and its answer (True or False) to ChatGPT, and collect reasoning analysis by following the prompt.

\textbf{Prompt:} \textit{Here is a question about law, could you please answer it and give me a detailed analysis?}

\textit{\#\#\# Question: }\texttt{[Question]}

\textit{\#\#\# Directions: The answer of given question can be "\texttt{[Yes/No]}" or "\texttt{[Correct/Incorrect]}". Please answer it step by step. If the question is about case analysis, please analyze the reasons behind the party's decision to take such action. If the question is about legal concept, please list the legal basis involved in your answer.}

\textit{Output:}

After examining the replies produced by ChatGPT, we find although the conclusions of most examples are correct, the reasoning analysis might be illogical, which refers to the wrong legal basis. To further improve the quality of the dataset, we collect the analysis of Judicial Examination questions in the past two years, which are written by human experts. We also ask experts to transfer the questions manually. Thanks to their efforts, we can obtain a high-quality dataset of 850 examples.

In Table~\ref{tab:fakao_version_examples}, we show examples of different methods of collecting explanations for JE questions. The answer in the JE-Q2EA example is wrong, let alone its explanation. 
For the JE-QA2E example, the answer to the question is correct, but the reasoning process is questionable, where a non-existent legal article.
The \textbf{JE-\textsc{Expert}} example is of the best quality, citing the correct law and performing logical reasoning.

\subsection{Legal Consultation}
\label{appendix:legal_consultation}
To generate the response to a legal consultation question, we ask ChatGPT to act as a lawyer and respond to a client.
In the input prompt, we list six requirements that the generated responses should meet: 
(1) properly citing legal articles, (2) giving well-founded analyses based on the facts of the case and legal articles, (3) responding comprehensively and analyzing the potential possibilities, (4) asking appropriate questions to dig out facts to assist in further answers, (5) using plain language, (6) giving preliminary legal opinions and consulting conclusions.

In the example of  Table~\ref{tab:article_multi_turn_en}, we retrieve 3 legal articles to facilitate a more faithful generation. Note that two of them are irrelevant to the consultation and ChatGPT is able to ignore them in its response.

\begin{table*}[ht]
\small
\centering
\begin{tabular}{p{0.8\columnwidth}p{1.2\columnwidth}}
\toprule
\zhsmall{\textbf{\textit{Related Articles:}} 《民法典》第一千零四十七条：结婚年龄，男不得早于二十二周岁，女不得早于二十周岁。
《民法典》第一千零四十六条：结婚应当男女双方完全自愿，禁止任何一方对另一方加以强迫，禁止任何组织或者个人加以干涉。
《民法典》第三百零四条：共有人可以协商确定分割方式。达不成协议...}                                                                              & \textbf{\textit{Related Articles:}} Article 1,047 of the Civil Code: A man shall attain the age of twenty-two, and a woman the age of twenty to enter into matrimony. Article 1,046 of the Civil Code: A man and a woman shall enter into matrimony freely and voluntarily. Neither party may compel the other to conclude a marriage against the latter’s will, and no organization or individual may interfere with the freedom of marriage. Article 304 of the Civil Code: Co-owners may determine the means of partition of their co-owned property by consultation. Where they fail to reach an agreement...
\\
\\

\zhsmall{\textbf{Client:} 我和我对象想结婚，我想问一下，结婚的法定婚龄是几岁啊？}                                                                              & \textbf{Client:} I would like to ask, what is the legal minimum age for marriage?  
\\
\\
\zhsmall{\textbf{Lawyer:} 根据《民法典》第一千零四十七条规定，结婚的法定婚龄为男不得早于22周岁，女不得早于20周岁。如果未到法定结婚年龄而坚持结婚，将被认定为无效婚姻。请问您和您的对象的年龄是否已达到法定婚龄？如果未达到，建议您等待至法定婚龄再考虑结婚的问题。} & \textbf{Lawyer:} According to Article 1,047 of the Civil Code, the legal marriageable age is a minimum of 22 years old for males and 20 years old for females. If individuals insist on getting married before reaching the legal marriageable age, the marriage will be deemed void. May I ask if both you and your partner have reached the legal marriageable age? If not, I would suggest waiting until you reach the legal marriageable age before considering marriage. \\
\bottomrule
\end{tabular}
\caption{An example of the retrieval augmented legal consultation generated by ChatGPT in Section~\ref{sec:legal consultation}.}
\label{tab:article_multi_turn_en}
\end{table*}

\section{Details of Charge Prediction Tasks}
\label{sec:legal_cla_details}

CAIL2018~\cite{Xiao2018CAIL2018AL} is large-scale dataset for Judgment Prediction, which is constructed from the criminal documents collected from China Judgments Online\footnote{\url{https://wenshu.court.gov.cn/}}. It includes three tasks, to predict the charge, to retrieve related articles, and to predict the terms of penalty. Since most criminal documents mention more than one articles, LLMs cannot give validate predictions zero-shot via perplexity. And previous also reveal that LLMs are not sensitive to numerical texts~\cite{chen-etal-2023-improving}. Thus, we do not choose \textit{Article Retrieval} and \textit{Prediction of Terms-of-Penalty} as the evaluation benchmark.

For Charge Prediction, we first filter out the documents which refer to multiple charges. And to construct a challenging task, we select several sets of charges which are easily misidentified to each other. The charges are:

(1) Forcible seizure~(\zhsmall{抢夺罪}) and Robbery~(\zhsmall{抢劫罪});

(2) Theft~(\zhsmall{盗窃罪}) and Criminal Conversion~(\zhsmall{侵占罪});

(3) Arsom~(\zhsmall{放火罪}) and Arson by Negligence~(\zhsmall{失火罪});

(4) Offering Bribes~(\zhsmall{行贿罪}), Acceptance of Bribes~(\zhsmall{受贿罪}) and Embezzlement~(\zhsmall{贪污罪}).

For each charge, we randomly sample 100 documents from CAIL2018. When evaluating, the model should choose a charge from the nine ones as its prediction. We employ zero-shot accuracy as the evaluating metric.

\section{Examples from Human Evaluation}
Here, we analyze some cases from the human evaluation.

In Table~\ref{tab:human_eval_examples_different_je_data}, we show the responses outputted by the models trained with different JE datasets.
In the output by the model trained with JE-Q2EA-42k, there is a misjudgment of the effectiveness of the agreement. The legal act of divorce will only invalidate the marriage relationship, but will not necessarily make the agreement lose the legal validity of the contract.
The model trained with JE-QA2E-6k misunderstands Article 1083 of the Civil Code and claims that \textit{if the husband wants to remarry after changing his mind, he can apply to the marriage registration authority to register a new marriage and establish a new marital relationship}. Remarriage requires both parties to jointly apply for re-marriage registration, and cannot be applied by one party alone.
The biggest advantage of the output by the model trained on \textsc{JE-Expert}, compared with the other two editions, is that the validity of the agreement is discussed under different circumstances. 
The parties did not provide detailed agreement content, but the model infers that there will be agreements on property division, child support, and possibly remarriage in the agreement, showing its reasoning ability. 
At the same time, the model further infers that remarriage is a voluntary act based on the provisions of the Civil Code and cannot be unilaterally forced to remarry, demonstrating the ability to reason according to the law.

In Table~\ref{tab:hallucination_examples}, we show an example where legal article retrieval helps alleviate the hallucinations of models.
Although the model has been trained on the corpus of legal articles, it cannot recall it accurately when needed. If we ask the model whether cousins can marry without providing relevant legal articles in the input, it will exhibit severe hallucinations: It quotes the wrong number of article and adds non-existent content after a real legal article.
When augmented with three relevant articles, the model will give a correct answer based on the correct article, with no hallucinations.

\begin{table*}[ht]
\small
\centering
\begin{tabular}{p{0.8\columnwidth}p{1.2\columnwidth}}
\toprule
\multicolumn{2}{c}{\textbf{JE-Q2EA}} \\
\midrule
\zhsmall{\textbf{\textit{Question:}} 关于抢劫罪，下列说法错误吗？\newline 驾驶摩托车夺取他人财物时，因被害人不放手而采用强拉硬拽方法劫取财物的，应认定为抢劫罪。\newline 请给出详细的推理过程之后再给出答案。}         & 
\small{\textbf{\textit{Question:}} Is the following statement about the crime of robbery incorrect? \newline
If someone drives a motorcycle and forcibly pulls and drags someone's property because the victim does not let go, it should be recognized as the crime of robbery. \newline
Please provide a detailed reasoning process before giving the answer.}
\\
\zhsmall{\textbf{Answer:} 这个说法是错误的。\newline根据《中华人民共和国刑法》第X条规定，抢劫罪是指以暴力、胁迫方法抢夺公私财物的行为。而强拉硬拽并不属于暴力、胁迫的范畴，因此不符合抢劫罪的构成要件。\newline此外，即使被害人不放手，也不能因此就认定为采用了暴力、胁迫的手段。因此，驾驶摩托车夺取他人财物时，因被害人不放手而采用强拉硬拽方法劫取财物的行为，应认定为盗窃罪。}                                                                              & \textbf{Answer:} This statement is incorrect.\newline According to Article X of the Criminal Law of the People's Republic of China, the crime of robbery refers to the act of forcibly seizing public or private property through violence or coercion. However, forcibly pulling and snatching does not fall within the scope of violence or coercion, and therefore does not meet the criminal elements of the crime of robbery. \newline 
Furthermore, even if the victim does not let go, it does not automatically imply the use of violence or coercion. Therefore, when someone drives a motorcycle and forcibly pulls and snatches someone's property because the victim does not let go, it should be recognized as the crime of theft.

\\
\midrule
\midrule
\multicolumn{2}{c}{\textbf{JE-QA2E}} \\
\midrule
\zhsmall{\textbf{\textit{Question:}} 关于抢劫罪，驾驶摩托车夺取他人财物时，因被害人不放手而采用强拉硬拽方法劫取财物的，应认定为抢劫罪。这个说法是否正确？}                                                               & 
\small{\textbf{\textit{Question:}} Is it correct to classify the act of using forceful pulling and dragging methods to seize someone's property while driving a motorcycle as robbery?}
\\

\zhsmall{\textbf{Answer:} 根据《中华人民共和国刑法》第二百六十四条规定，抢劫罪是指以暴力、胁迫方法抢夺公私财物的行为。而在本题中，驾驶摩托车夺取他人财物时，因被害人不放手而采用强拉硬拽方法劫取财物，显然是采用了暴力、胁迫手段，因此符合抢劫罪的构成要件。此外，根据《中华人民共和国刑法》第六十六条规定，使用暴力、胁迫方法抢夺公私财物，情节严重的，应当判处三年以上、十年以下有期徒刑，并处罚金或者没收财产。因此，对于这里的行为，应当认定为抢劫罪，并依法追究刑事责任。综上所述，这里的正确回答为“正确”，因为该行为符合抢劫罪的构成要件，根据相关法律规定应当认定为抢劫罪。}                                                                              & \textbf{Answer:} According to Article 264 of the Criminal Law of the People's Republic of China, robbery refers to the act of forcibly taking public or private property through violence or coercion. In this case, when driving a motorcycle to seize someone's property and resorting to forceful pulling and dragging because the victim refuses to let go, it is evident that violent and coercive means are being employed. Therefore, it meets the criminal elements of robbery. Furthermore, according to Article 66 of the Criminal Law of the People's Republic of China, if violence or coercion is used to commit robbery and the circumstances are serious, the offender shall be sentenced to imprisonment for more than three years but less than ten years, in addition to a fine or confiscation of property. Therefore, for this behavior, it should be classified as robbery and the criminal responsibility should be pursued in accordance with the law. In conclusion, the correct answer to this question is "correct" because this behavior meets the criminal elements of robbery and should be classified as such according to relevant legal provisions.
\\
\midrule
\midrule
\multicolumn{2}{c}{\textbf{\textsc{JE-Expert}}} \\
\midrule
\zhsmall{\textbf{\textit{Question:}} 关于抢劫罪，驾驶摩托车夺取他人财物时，因被害人不放手而采用强拉硬拽方法劫取财物的，应认定为抢劫罪。这个说法是否正确？}                                                               & 
\small{\textbf{\textit{Question:}} Is it correct to classify the act of using forceful pulling and dragging methods to seize someone's property while driving a motorcycle as robbery?}
\\

\zhsmall{\textbf{Answer:} 在刑法理论上，构成抢劫罪的构成要件之一是行为人对被害人当场使用暴力、胁迫或者其他方法。如果犯罪分子没有使用暴力或者胁迫的方法就取得了财物，不能以抢劫罪论处。但是，如果犯罪分子事先只是准备抢夺,但在实施抢夺的过程中遭到反抗或者阻拦,于是当场使用暴力或者以暴力相威胁强取财物,其行为就由盗窃或者抢夺转化为抢劫了,应以抢劫罪定罪处罚。在实体法上，根据《最高人民法院关于审理抢劫、抢夺刑事案件适用法律若干问题的意见》:"驾驶车辆强抢财物时，因被害人不放手而采取强拉硬拽方法劫取财物的，应当以抢劫罪定罪处罚。"因此，本案定抢劫罪的说法正确。}                                                                              & \textbf{Answer:} In the theoretical framework of criminal law, one of the criminal elements of the crime of robbery is that the perpetrator employs violence, coercion, or other methods against the victim on the spot. If the criminal obtains property without using violence or coercion, they cannot be charged with robbery. However, if the criminal only intended to commit robbery beforehand but encounters resistance or obstruction during the act of robbery and subsequently employs violence or threatens to use violence to forcibly take the property, their behavior transforms from theft or robbery to robbery, and they should be convicted and punished for the crime of robbery. In terms of substantive law, according to the "Opinions of the Supreme People's Court on Several Issues concerning the Application of Laws in the Trial of Robbery and Extortion Criminal Cases": "When a person drives a vehicle and forcibly snatches property, if they resort to forcibly pulling and dragging to seize the property because the victim does not let go, they should be convicted and punished for the crime of robbery." Therefore, the claim that this case falls under the crime of robbery is correct.
\\
\bottomrule
\end{tabular}
\caption{Examples of the three different methods for collecting JE explanations in Section~\ref{sec:National Judicial Examination}.}
\label{tab:fakao_version_examples}
\end{table*}

\begin{table*}[ht]
\small
\centering
\begin{tabular}{p{0.4\columnwidth}p{0.6\columnwidth}p{0.4\columnwidth}p{0.6\columnwidth}}

\toprule
\zhsmall{请问：民法典中关于“婚姻无效”是如何规定的？ }   & \small{May I ask: How is "void marriage" defined in the Civil Code?} & \zhsmall{没领证可以生小孩吗？需要办什么手续？} & \small{Can we have a baby without a marriage certificate? What procedures are required?}\\
\midrule      
\zhsmall{大概一年多前，父母帮我们订了婚约，但现在一点感觉都没有了，我能不和他结婚吗？婚约一定得遵守吗？  }                 & \small{Approximately over a year ago, my parents helped us arrange an engagement, but now I don't have any feelings at all. Can I not marry him? Do engagements have to be honored?}&  \zhsmall{2022年，甲因胁迫结婚，请求撤销婚姻，应当自胁迫行为终止之日起两年内提出，对吗？}      & \small{In 2022, Bob was coerced into marriage and now requests the annulment of the marriage. The request should be made within two years from the date of termination of the coercion, right?}  \\
\midrule      

\zhsmall{邻居两口子为了买房子办了假离婚，还签了协议来着，结果现在女的反悔了，不复婚了。男的不干了，啥也没有了，他们签的这协议有法律效力吗？} & \small{The neighbors, a married couple, got a fake divorce to buy a house, and they even signed an agreement. But now the woman has changed her mind and doesn't want to remarry. The man disagrees because he has nothing left. Does the agreement they signed have legal validity?}  & \zhsmall{实在受不了他抽烟喝酒的，还是离了好。也没孩子，其他都谈好了，就差房子了，反正我是没有其他地方住了，这房子我必须要。他也想要房子，怎么办呢？} & \small{I can't stand his smoking and drinking anymore. It's better to get a divorce. We don't have any children, and everything else is settled, except for the house. Anyway, I have nowhere else to live, so I must have this house. He also wants the house, so what should we do?}  
\\
\midrule      

\zhsmall{男方婚内出轨，还和小三生了一个孩子，该怎么办？能让他净身出户吗？  }                                & \small{How to deal with it when the husband cheats during the marriage and has a child with the other woman? Can he leave without taking any property after divorce?}& \zhsmall{终于决定要分居了，虽然还没走到离婚那一步。分居协议书该怎么写呢？} & \small{Finally decided to separate, although we haven't reached the divorce stage yet. How should we write a separation agreement?}
\\
\midrule      

\zhsmall{离婚冷静期是什么意思？}  & \small{What is divorce cooling-off period?} & \zhsmall{表兄妹可以结婚吗？}  & \small{Can cousins marry?}  \\
\midrule      
\zhsmall{不给孩子的抚养费最严重的后果是什么？}    & \small{What are the most serious consequences of not providing child support?}  & \zhsmall{我们俩已经分句两年了，算自动离婚吗？}                                                    & \small{We have been living separately for two years. Does that count as an automatic divorce?}      \\
\midrule      
\zhsmall{全职母亲离婚时可以分财产吗？}  & \small{Can a full-time mother get a share of the property in a divorce?}  & \zhsmall{彩礼是夫妻共同财产吗？}   & \small{Is bride price considered joint property of the husband and wife?} \\
\midrule  
\zhsmall{结婚前我爸妈给全款买的房，在城西，涨了一倍还多，现在如果离婚的话，要给她分一半吗？}                         & \small{Before getting married, my parents bought the house with full payment in the western part of the city. It has increased by more than double in value. Now, if we were to divorce, would I have to give her half?}  & \zhsmall{大概5,6年前，我们俩就住一起了，一直也没领证，也没要孩子。现在他老打我，实在受不了了，我能去法院起诉离婚吗？}    & \small{Approximately 5 or 6 years ago, the two of us started living together, but we never obtained marriage certificate or had children. Now he constantly abuses me, and I can't tolerate it anymore. Can I go to court and file for divorce?}                                           \\
\midrule      
\zhsmall{结婚证丢了，我还能去法院起诉他离婚吗？}            & \small{Can I still sue him for divorce in court if I lost my marriage certificate?} & \zhsmall{女婿有没有赡养岳父的义务}  & \small{Does the son-in-law have an obligation to support his father-in-law?}   
\\
\midrule      
\zhsmall{孩子还不满一周，能离婚吗？}      & \small{The child is not even a week old. Can we get a divorce?}& \zhsmall{总是被家暴，该怎么保护自己啊？}                & \small{I am constantly experiencing domestic violence. How can I protect myself?}    
\\
\bottomrule
\end{tabular}
\caption{List of the questions for human evaluation. (Part 1)}
\label{tab:eval_questions1}
\end{table*}

\begin{table*}[ht]
\small
\centering
\begin{tabular}{p{0.4\columnwidth}p{0.6\columnwidth}p{0.4\columnwidth}p{0.6\columnwidth}}
\toprule
\zhsmall{继父和我妈已经办了离婚手续了，我对他还有赡养义务吗？还要给他抚养费吗？ }   & \small{Stepfather and my mother have already completed the divorce proceedings. Do I still have an obligation to support him? Do I still need to pay alimony to him?}  & \zhsmall{她结婚前也没说自己得了红斑狼疮啊，能离婚吗？} & \small{She didn't mention having lupus erythematosus before getting married. Can we get a divorce?} \\
\midrule      

\zhsmall{我们俩本来都要结婚了，结果因为彩礼钱黄了，她爸妈又要5万分手费，这合法吗？} & \small{We were both planning to get married, but things fell through because of the bride price. Her parents are now demanding 50,000 yuan as a breakup fee. Is this legal?} & \zhsmall{结婚以后，我爸妈出钱买的房子，只写了我一个人的名字，这算是夫妻共同财产吗？} & \small{After getting married, my parents bought a house and only registered my name on it. Is it community property?}\\
\midrule      
 \zhsmall{我自己都吃不上饭了，还要给养父赡养费吗？}  & \small{I can't even afford to feed myself. Do I still have to pay support to my foster father?}  
& \zhsmall{我是朝鲜族人，今年虚岁20了，可以领结婚证吗？} & \small{I am of Korean ethnicity, and I am 20 years old by traditional reckoning. Can I apply for a marriage certificate?}
\\
\midrule      
\zhsmall{复婚和再婚有什么区别吗？}  & \small{What is the difference between remarriage and second marriage?} & \zhsmall{同姓的人可以结婚吗？}  & \small{Can people with the same surname get married?}  \\
\midrule      
\zhsmall{领养小孩需要啥条件？}    & \small{What are the requirements for adopting a child?}  & \zhsmall{军人多少岁可以结婚？}                                & \small{At what age can military personnel get married?}    \\
\midrule      
\zhsmall{30岁领证算晚婚吗？}                         & \small{Is getting married at 30 considered a late marriage?}  & \zhsmall{离婚多久能复婚？}   & \small{How long after a divorce can you remarry?} \\
\midrule  
\zhsmall{刚离婚就去登记算重婚吗？}   & \small{it considered bigamy if you register another marriage immediately after getting divorced?} & \zhsmall{离异再婚要改户口本吗？}    & \small{Do you need to update the household registeration for remarriage after divorce?}                                           \\
\midrule      
\zhsmall{再婚的时候怎么保护婚前个人财产}            & \small{How can I protect my personal property acquired before marriage when remarrying?} & \zhsmall{没有户口本能领结婚证吗？}  & \small{can you get a marriage certificate without a household registration?}   
\\
\midrule      
\zhsmall{结婚证丢了怎么补办？}            & \small{What should I do if I lost my marriage certificate and need to renew it?} &   & 
\\
\bottomrule                                        
\end{tabular}
\caption{List of the questions for human evaluation. (Part 2)}
\label{tab:eval_questions2}
\end{table*}

\begin{table*}[ht]
\small
\centering
\begin{tabular}{p{0.8\columnwidth}p{1.2\columnwidth}}
\toprule
\zhsmall{\textbf{Question:} 邻居两口子为了买房子办了假离婚，还签了协议来着，结果现在女的反悔了，不复婚了。男的不干了，啥也没有了，他们签的这协议有法律效力吗？}                                                                              & \textbf{Question:} The neighbors, a married couple, got a fake divorce to buy a house, and they even signed an agreement. But now the woman has changed her mind and doesn't want to remarry. The man disagrees because he has nothing left. Does the agreement they signed have legal validity?
\\
\midrule
\zhsmall{\textbf{Articles:} 《民法典》第一千零八十条：完成离婚登记，或者离婚判决书、调解书生效，即解除婚姻关系。\newline
《民法典》第一千零八十三条：离婚后，男女双方自愿恢复婚姻关系的，应当到婚姻登记机关重新进行结婚登记。\newline 
《民法典》第一千零九十八条：收养人应当同时具备下列条件：（一）无子女或者只有一名子女；（二）有抚养、教育和保护被收养人的能力；（三）未患有在医学上认为不应当收养子女的疾病；（四）无不利于被收养人健康成长的违法犯罪记录；（五）年满三十周岁。
} & \textbf{Articles:} Article 1,080 of the Civil Code: A marital relationship is dissolved on completion of registration of divorce or when a judgment granting divorce or a mediation agreement on divorce takes effect. \newline
Article 1,083 of the Civil Code: Where, after divorce, both the man and the woman voluntarily resume their marital relationship, they shall once again file for registration of marriage 
at a marriage registry.\newline
Article 1,098 of the Civil Code: A prospective adopter shall meet all of the following criteria:
1. Having no child or only one child;
2. Being capable of raising, educating and protecting the adoptee;
3. Being free of any disease which renders the prospective adopter 
medically unfit to be an adopter; 
4. Having no criminal record unfavorable to the healthy growth of the adoptee; and 
5. Having attained the age of thirty. 

 \\
 \midrule
\zhsmall{\textbf{\textsc{JE-Expert}: }您好，根据《民法典》的规定，完成离婚登记或者离婚判决书、调解书生效，即解除婚姻关系。因此，如果邻居两口子已经办理了离婚登记手续，那么他们的婚姻关系已经解除。如果他们签署了协议，该协议是否具有法律效力需要看具体情况。\newline首先，如果该协议是关于离婚后的财产分割、赡养子女等方面的约定，那么该协议是有效的。因为夫妻离婚后，财产分割、赡养子女等问题需要通过协商达成一致意见，并签订书面协议。如果没有违反法律规定，该协议是具有法律效力的。\newline其次，如果该协议是关于复婚方面的约定，那么该协议不一定有效。因为复婚是夫妻自愿的行为，如果女方不同意复婚，男方单方面无法强制复婚。\newline最后，建议您了解一下具体的情况，比如该协议的具体内容、签订时间以及是否符合相关法律法规的规定等。如果您需要进一步的帮助，可以向当地的律师咨询。} 
& \small{\textbf{\textsc{JE-Expert}: }Hello, according to the provisions of the Civil Code, completing the registration of divorce or the effective divorce judgment or mediation agreement will dissolve the marriage relationship. Therefore, if your neighbors have already completed the divorce registration procedures, their marriage relationship has been dissolved. Whether their agreement has legal effect depends on the specific circumstances.\newline
Firstly, if the agreement pertains to the division of property and child support after divorce, then the agreement is valid. After a couple divorces, issues such as property division and child support need to be agreed upon through negotiation and documented in a written agreement. If the agreement does not violate any legal provisions, it has legal effect. \newline 
Secondly, if the agreement pertains to remarriage, then the agreement may not be valid. Remarriage is a voluntary act between spouses, and if one party does not agree to remarry, the other party cannot force them to do so unilaterally. \newline 
Lastly, it is advisable to gather specific information about the situation, such as the exact content of the agreement, the date of signing, and whether it complies with relevant laws and regulations. If you need further assistance, you can consult a local lawyer.}
\\ 
\midrule
\zhsmall{\textbf{JE-Q2EA-42k: }您好，根据《民法典》第一千零八十条规定，完成离婚登记或者离婚判决书、调解书生效，即解除婚姻关系。因此，如果两人已经办理了离婚手续并签署了协议，那么这个协议就无效了。\newline但是，如果这个协议是关于房屋买卖的相关事宜，可以根据合同法的规定来解决。请问这个协议的具体内容是什么？是否涉及到房屋买卖等相关事宜？} 
& 
\small{\textbf{JE-Q2EA-42k: }Hello, according to Article 1080 of the Civil Code, the divorce registration, divorce judgment, or mediation agreement becomes effective and terminates the marital relationship. Therefore, if both parties have completed the divorce procedures and signed an agreement, that agreement is considered invalid.\newline
However, if the agreement pertains to matters related to property transactions, such as the sale of a house, it can be resolved in accordance with the provisions of the Contract Law. Could you please provide the specific content of this agreement? Does it involve matters related to the sale of a house or other related issues?
}
\\
\midrule
\zhsmall{\textbf{JE-QA2E-6k: }您好，根据《民法典》第一千零八十条规定，完成离婚登记，或者离婚判决书、调解书生效，即解除婚姻关系。因此，如果两人已经办理了离婚手续并签署了离婚协议，那么这份协议就生效了，具有法律效力。但是如果男方反悔想要复婚，那么他可以向婚姻登记机关申请重新进行结婚登记，重新建立婚姻关系。\newline 请问你们之间是否已经完成离婚登记呢？如果没有完成离婚登记，那么这份协议就没有生效，也就不存在法律效力。} & 
\small{\textbf{JE-QA2E-6k: }Hello, according to Article 1080 of the Civil Code, completing the divorce registration or having the divorce judgment or mediation agreement take effect will terminate the marital relationship. Therefore, if both parties have already gone through the divorce procedures and signed the divorce agreement, then this agreement is valid and has legal effect. However, if the husband wants to remarry after changing his mind, he can apply to the marriage registration authority to register a new marriage and establish a new marital relationship.\newline
May I ask if you have already completed the divorce registration between you? If the divorce registration has not been completed, then this agreement is not effective and does not have legal force.
}
\\
\bottomrule
\end{tabular}
\caption{Examples of the responses outputted by the models trained with different JE datasets in Section~\ref{sec:sft_quality}.}
\label{tab:human_eval_examples_different_je_data}
\end{table*}

\begin{table*}[ht]
\small
\centering
\begin{tabular}{p{0.8\columnwidth}p{1.2\columnwidth}}
\toprule
\zhsmall{\textbf{Question:} 表兄妹可以结婚吗？}                                                                              & \textbf{Question:} Can cousins marry?
\\
\midrule
\zhsmall{\textbf{Articles:}《民法典》第一千零四十八条：直系血亲或者三代以内的旁系血亲禁止结婚。\newline
《民法典》第一千零五十一条：有下列情形之一的，婚姻无效：（一）重婚；（二）有禁止结婚的亲属关系；（三）未到法定婚龄。\newline
《民法典》第一千零四十九条：要求结婚的男女双方应当亲自到婚姻登记机关申请结婚登记。符合本法规定的，予以登记，发给结婚证。完成结婚登记，即确立婚姻关系。未办理结婚登记的，应当补办登记。} & \textbf{Articles:} Article 1048 of the Civil Code: Persons who are lineal relatives by blood, or collateral relatives by blood within three generations are prohibited from marrying each other.\newline
Article 1051 of the Civil Code: A marriage is null and void under any of the following circumstances:
1. Bigamy;
2. The parties to the marriage are within the prohibited degrees of relationship; or
3. Either party to the marriage has not yet attained the legal marriage age.\newline
Article 1049 of the Civil Code: Both the man and the woman intending to enter into matrimony shall apply for registration of the marriage in person with a marriage registry. If the marriage proposed is found to conform to the provisions of this Code, the marriage shall be registered, and a marriage certificate shall be issued. A marital relationship shall form on the completion of the marriage registration. A couple who has failed to file a marriage registration shall complete the registration. 
 \\
 \midrule
\textbf{w/ Hallucination:} \zhsmall{《婚姻法》第二十一条第一款规定：“有下列情形之一的，为无效婚姻：（一）重婚的；（二）有禁止结婚的亲属关系的；（三）未到法定婚龄的；（四）被法律认定为无民事行为能力人或者限制民事行为能力人的；（五）在婚姻登记过程中弄虚作假，骗取婚姻登记的；（六）其他违法婚姻的。”据此，表兄妹是三代以内旁系血亲，属于《民法典》禁止结婚的亲属关系，因此，他们不能结婚。}
& 
\textbf{w/ Hallucination:}
Article 21, paragraph 1 of the Marriage Law states: "A marriage shall be void under any of the following circumstances: (1) bigamy; (2) marriage between relatives within the prohibited degrees; (3) marriage before reaching the legal marriage age; (4) marriage involving a person who has been recognized by law as lacking capacity for civil conduct or whose capacity for civil conduct is restricted; (5) obtaining marriage registration through fraudulent means during the marriage registration process; (6) other illegal marriages." Based on this, first cousins are blood relatives within the third degree of collateral consanguinity, which falls under the prohibited degrees of marriage according to the Civil Code. Therefore, they cannot get married.
\\
 \midrule
\textbf{w/o Hallucination:}  \zhsmall{表兄妹是三代以内的旁系血亲，因此是不能在民法典范围内确定婚姻关系的。} 
& 
\textbf{w/o Hallucination:} Cousins are collateral blood relatives within three generations, therefore their marriage relationship cannot be established within the scope of the Civil Code.
\\ 
\bottomrule
\end{tabular}
\caption{An example of (1) the model output with hallucinations when not provided with legal articles and (2) the model output without hallucinations when augmented with retrieved legal articles in Section~\ref{sec:retrieval_alleviates_hallucinaiton}. }
\label{tab:hallucination_examples}
\end{table*}

\end{document}